%% file: main.tex
\documentclass[runningheads]{llncs}
\usepackage{graphicx}

\usepackage{tikz}
\usepackage{comment}
\usepackage{amsmath,amssymb} 
\usepackage{color}

\usepackage{xspace}
\usepackage{graphicx}
\usepackage{booktabs}

\usepackage{hyperref}

\newcommand*{\ie}{i.e.\@\xspace}
\newcommand*{\eg}{e.g.\@\xspace}

\usepackage[capitalize]{cleveref}
\crefname{section}{Sec.}{Secs.}
\Crefname{section}{Section}{Sections}
\Crefname{table}{Table}{Tables}
\crefname{table}{Tab.}{Tabs.}

\begin{document}
\pagestyle{headings}
\mainmatter
\def\ECCVSubNumber{7044}  

\title{On Label Granularity and Object Localization} 

\titlerunning{On Label Granularity and Object Localization}
%

\author{
{Elijah Cole\inst{1,\dagger} \quad Kimberly Wilber\inst{2} \quad Grant Van Horn\index{Van Horn, Grant}\inst{3} \quad Xuan Yang\inst{2}} \\
{Marco Fornoni\inst{2} \quad Pietro Perona\inst{1} \quad Serge Belongie\inst{4}} \\
{Andrew Howard\inst{2} \quad  Oisin Mac Aodha\index{Mac Aodha, Oisin}\inst{5} }
}

\authorrunning{E. Cole et al.}
%
\institute{$^1$Caltech \quad $^2$Google \quad $^3$Cornell University \\
$^4$University of Copenhagen \quad $^{5}$University of Edinburgh\\~\\ $^\dagger$\href{mailto:ecole@caltech.com}{ecole@caltech.edu}}
\maketitle

\begin{abstract}
Weakly supervised object localization (WSOL) aims to learn representations that encode object location using only image-level category labels. However, many objects can be labeled at different levels of granularity. Is it an animal, a bird, or a great horned owl? Which image-level labels should we use? In this paper we study the role of label granularity in WSOL. To facilitate this investigation we introduce iNatLoc500, a new large-scale fine-grained benchmark dataset for WSOL. Surprisingly, we find that choosing the right training label granularity provides a much larger performance boost than choosing the best WSOL algorithm. We also show that changing the label granularity can significantly improve data efficiency. 
\end{abstract}

\section{Introduction}

For many problems in computer vision, it is not enough to know \emph{what} is in an image, we also need to know \emph{where} it is. Examples can be found in many domains, including ecological conservation \cite{van2014nature}, autonomous driving \cite{wu2017squeezedet}, and medical image analysis \cite{li2019clu}. The most popular paradigm for locating objects in images is object \emph{detection}, which aims to predict a bounding box for every instance of every category of interest. Object \emph{localization} is special case of detection where each image is assumed to contain exactly one object instance of interest, and the category of that object is known. 

\begin{figure}[htb]
\centering
\includegraphics[width=0.9\linewidth]{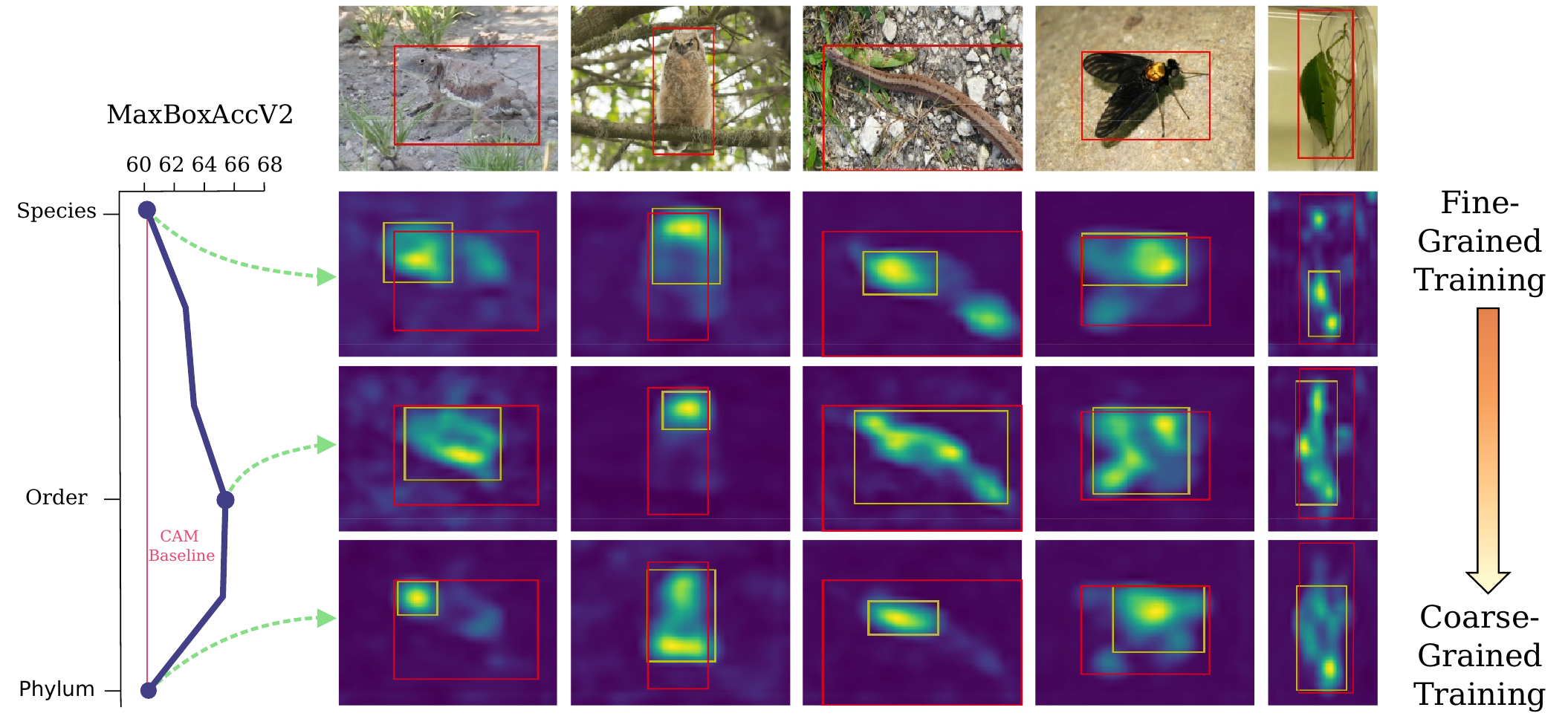}
\caption{
\textbf{Label granularity is a critical but understudied factor in weakly supervised object localization~(WSOL).}
We show five hand-picked examples from our iNatLoc500 dataset. Below each image we show class activation maps (CAMs)~\cite{zhou2016learning} derived from training a classifier at different granularity levels, with ground truth bounding boxes (red) and WSOL-based bounding boxes (yellow) superimposed. Conventional training does not consider label granularity and can lead to inferior localization performance (red line). Better WSOL results can be achieved by training with coarse (\ie ``order") labels, as opposed to fine-grained (\ie ``species") ones. 
}
\label{fig:teaser}
\end{figure}

Standard approaches to object detection and localization require bounding boxes for training, which are expensive to collect at scale~\cite{papadopoulos2017extreme}. Weakly supervised object localization (WSOL) methods aim to sidestep this obstacle by learning to localize objects using only image-level labels at training time. The potential reduction in annotation cost which could result from effective weakly supervised methods has stimulated significant interest in WSOL over the last few years~\cite{zhang2021weakly}.

In this paper we explore the role of label granularity in WSOL. The \emph{granularity} of a category is the degree to which it is specific, which can vary from coarse-grained (\eg ``animal") to fine-grained (\eg ``great horned owl")~\cite{wei2021fine}. When we work with benchmark datasets in computer vision, we often take the given level of label granularity for granted. However, it is usually possible to make those labels more general or more specific. It is worth asking whether the label granularity we are given is the best one to use for a certain task. Label granularity matters for WSOL because the first step in most WSOL algorithms is to train a classifier using image-level category labels. By choosing a label granularity we are choosing which training images are grouped into categories. This affects the discriminative features learned by the classifier and ultimately determines the bounding box predictions. Is it possible to improve WSOL performance by controlling label granularity? 

Unfortunately, it is difficult to explore label granularity in WSOL due to the limitations of existing datasets. The field of WSOL largely relies on CUB~\cite{WahCUB_200_2011} and ImageNet~\cite{russakovsky2015imagenet}. CUB has a consistent label hierarchy (\ie one that can be used to measure label granularity), but it is small ($\sim$6k training images) and homogeneous (only bird categories). ImageNet is large and diverse, but lacks a consistent label hierarchy (see Sec.~\ref{sec:dataset_props}). Furthermore, \cite{choe2020evaluating} recently found that many purported algorithmic advances in WSOL over the last few years -- which were based on these two datasets -- perform no better than baselines when they are evaluated fairly. This calls for the development of more diverse and challenging benchmarks for WSOL. 

\noindent
Our primary contributions are as follows:
\begin{enumerate}
    \item We explore the effect of label granularity on WSOL, and show that training at coarser levels of granularity leads to surprisingly large performance gains across many different WSOL methods compared to conventional training \eg +5.1 \texttt{MaxBoxAccV2} for CAM and +6.6 \texttt{MaxBoxAccV2} for CutMix (see Fig.~\ref{fig:granularity_wsol})
    \item We demonstrate that training on coarse labels is more data efficient than conventional training. For instance, training at a coarser level achieves the same performance as conventional CAM with $\sim 15\times$ fewer labels (see Fig.~\ref{fig:inatloc_weaksup}).
    \item We introduce the iNaturalist Localization 500 (iNatLoc500) dataset, which consists of 138k images for weakly supervised training and 25k images with manually verified bounding boxes for validation and testing. iNatLoc500 covers 500 diverse categories with a consistent hierarchical label space. 
\end{enumerate} 

\section{Related Work}
Here we primarily focus on literature related to WSOL. See~\cite{zhang2021weakly} for a broader overview of related techniques such as weakly supervised object detection~\cite{bilen2015weakly,bilen2016weakly,tang2017multiple}. 

\noindent{\bf Weakly Supervised Object Localization.} 
The goal of WSOL is to determine the location of single objects in images using only image-level labels at training time. Early attempts at WSOL explored a variety of different approaches, such as adapting boosting-based methods~\cite{opelt2005object}, framing the problem as multiple instance learning~\cite{galleguillos2008weakly,gokberk2014multi}, and applying latent deformable part-based formulations~\cite{pandey2011scene}. 

Some foundational work in deep learning investigated the degree to which object localization comes ``for free" when training supervised CNNs for image classification tasks~\cite{zeiler2014visualizing,oquab2015object,zhou2016learning}. In particular, the Class Activation Mapping (CAM) method of~\cite{zhou2016learning} showed that CNNs can capture some object location information even when they are trained using only image-level class labels. This inspired a large body of work (\eg~\cite{zhang2018adversarial,singh2017hide,zhang2018self,choe2019attention,ki2020sample,kim2021keep}) that attempted to address some of the shortcomings of CAM, \eg by preventing the underlying model from only focusing on the most discriminative parts of an object~\cite{yun2019cutmix} or increasing the spatial resolution of its outputs~\cite{selvaraju2017grad,chattopadhay2018grad}. 

Recently, \cite{choe2020evaluating} showed that when state-of-the-art WSOL methods are fairly compared (\eg by controlling for the backbone architecture and operating thresholds), they are no better than the standard CAM~\cite{zhou2016learning} baseline. Thus, despite its simplicity, CAM is still a surprisingly effective baseline for WSOL. Subsequent work has explored further techniques for improving CAM-based methods~\cite{bae2020rethinking,kim2021normalization} and alternative approaches for estimating model coefficients~\cite{jung2021towards}. 

\noindent{\bf Task Granularity and Localization.}
Despite the considerable interest in WSOL in recent years, many open questions remain. Examples include the effect of label granularity (\eg coarse-grained labels like ``bird" vs. fine-grained labels indicating the specific species of bird) and the effect of training set size. In the context of supervised object detection, \cite{van2018inaturalist} showed that \emph{coarsening} category labels at training time can improve the localization performance of \emph{object detectors}. It is unclear if the same phenomenon holds for WSOL. \cite{wang2020leads} explored the impact of label granularity for object detection on the OpenImages~\cite{kuznetsova2020open} dataset and observed a small performance improvement when training on finer labels. In the semi-supervised detection setting, \cite{yang2019detecting} trained object detectors on OpenImages and ImageNet using both coarse-grained bounding box annotations and fine-grained image-level labels. \cite{uijlings2018revisiting} also explored semi-supervised detection with an approach that generates object proposals across multiple hierarchical levels. Unlike our work, these detection-based methods require bounding box information at training time. In addition, the label hierarchies for datasets like ImageNet and OpenImages are not necessarily good proxies for visual similarity or concept granularity (see Sec.~\ref{sec:dataset_props}). 

For WSOL, \cite{kim2021keep} showed that aggregating class attribution maps at coarser hierarchical levels (\eg ``dog") results in more spatial coverage of the objects of interest, whereas maps for finer-scale concepts (\eg ``Afghan hound") only focus on subparts of the object. However, their analysis does not explore the impact of training at different granularity levels. It is also worth noting that their aggregation method only improves performance on CUB. Regarding data quantity, \cite{choe2020evaluating} studied the number of supervised examples used to tune the hyperparameters of CAM, but did not consider the impact of the number of examples used to train the image classifier. 

Though not directly related to our work, we note that label granularity has been studied in many contexts other than object localization, including action recognition \cite{shao2020finegym}, knowledge tracing \cite{choi2020ednet}, animal face alignment \cite{khan2020animalweb}, and fashion attribute recognition \cite{guo2019imaterialist}. In the context of image classification, prior work has tackled topics like analyzing the emergence of hierarchical structure in trained classifiers~\cite{bilal2017convolutional}, identifying patterns in visual concept generalization \cite{sariyildiz2021concept}, and training finer-grained image classifiers using only coarse-grained labels~\cite{taherkhani2019weakly,robinson2020strength,xu2021weakly,touvron2021grafit}.

\noindent{\bf Datasets for Object Localization.} 
Early work in WSOL (\eg \cite{opelt2005object,galleguillos2008weakly,nguyen2009weakly}) focused on relatively simple and small-scale datasets such as Caltech4~\cite{fergus2003object}, the Weizmann Horse Database~\cite{borenstein2002class}, or subsets of PASCAL-VOC~\cite{everingham2010pascal}. With the rise of deep learning-based methods, CUB~\cite{WahCUB_200_2011} and ImageNet~\cite{deng2009imagenet,russakovsky2015imagenet} became the standard benchmarks for this task. CUB~\cite{WahCUB_200_2011} consists of images of 200 different categories of birds, where each image contains a single bird instance. ImageNet~\cite{deng2009imagenet,russakovsky2015imagenet} contains 1000 diverse categories and has significantly more images than CUB ($>$1M compared to $\sim$6k). \cite{choe2020evaluating} proposed OpenImages30k, a 100-category localization-focused subset of the OpenImages V5 dataset~\cite{kuznetsova2020open}. An overview of these datasets is presented in Table~\ref{tab:datasets_overview}.  

These existing datasets are valuable, but they have shortcomings. CUB is small and homogeneous (only birds). OpenImages30k, as presented in \cite{choe2020evaluating}, is not actually evaluated as a bounding box localization task. It is instead a per-pixel foreground object segmentation task where the ground truth also features some ``ignore" regions that are excluded from the evaluation. Finally, while both OpenImages30k and ImageNet have label hierarchies, they do not reflect concept granularity in a consistent way. As a result, it is difficult to use them to better understand the relationship between concept granularity and localization. We discuss these issues in greater detail in Sec.~\ref{sec:dataset_props}. To address these shortcomings we introduce iNatLoc500, a new WSOL dataset composed of images from 500 fine-grained visual categories and equipped with a consistent label hierarchy. 

\begin{table*}[t]
\caption{
Comparison of datasets for WSOL. 
The vast majority of WSOL papers use only CUB and ImageNet. 
The OpenImages30k dataset was introduced by \cite{choe2020evaluating}, which also defines the splits we use for CUB and ImageNet. 
For each split we provide the minimum, maximum, and mean number of images per category, along with the total number of images in the split. 
Means are rounded to the nearest integer. 
The properties of these four datasets are discussed in detail in Sec.~\ref{sec:dataset_props}.
}
\centering
\resizebox{1.0\linewidth}{!}{
\begin{tabular}{|l|c|c|c|c|c|c|c|c|c|c|c|c|c|} \hline
& & \multicolumn{4}{c|}{\texttt{train-weaksup} ($D_w$)} & \multicolumn{4}{c|}{\texttt{train-fullsup} ($D_f$)} & \multicolumn{4}{c|}{\texttt{test} ($D_\mathrm{test}$)} \\
Dataset & \# Cat. & Min & Max & Mean & Total & Min & Max & Mean & Total & Min & Max & Mean & Total \\ 
\hline
CUB \cite{WahCUB_200_2011} & 200 & 29 & 30 & 30 & 6k & 3 & 6 & 5 & 1k & 11 & 30 & 29 & 5.8k \\
ImageNet \cite{deng2009imagenet} & 1000 & 732 & 1300 & 1281 & 1.28M & 10 & 10 & 10 & 10k & 10 & 10 & 10 & 10k\\
OpenImages30k \cite{benenson2019large,choe2020evaluating} & 100 & 230 & 300 & 298 & 30k & 25 & 25 & 25 & 2.5k & 50 & 50 & 50 & 5k \\
\hline
iNatLoc500 & 500 & 149 & 307 & 276 & 138k & 25 & 25 & 25 & 12.5k & 25 & 25 & 25 & 12.5k \\
\hline
\end{tabular}
}
\label{tab:datasets_overview}
\end{table*}

\section{Background}
\label{sec:background}

\subsection{Weakly Supervised Object Localization (WSOL)}
\label{sec:background_wsol}
We begin by formalizing the WSOL setting. 
Let $D_w$ be a set of \emph{weakly labeled} images, \ie $D_w = \{(x_i, y_i)\}_{i=1}^{N_w}$ where $x_i \in \mathbb{R}^{H \times W \times 3}$ is an image and $y_i \in \{1, \ldots, C\}$ is an image-level label corresponding to one of $C$ categories. Let $D_f$ be a set of \emph{fully labeled} images, \ie $D_f = \{(x_i, y_i, \textbf{b}_i)\}_{i=1}^{N_f}$ where $x_i$ and $y_i$ are defined as before and $\textbf{b}_i \in \mathbb{R}^4$ is a bounding box for an instance of category $y_i$. In practice $N_w \gg N_f$. WSOL approaches typically comprise three steps: 

\noindent
\textbf{(1) Train.} Use $D_w$ to train an image classifier $h_\theta:~\mathbb{R}^{H \times W \times 3}~\to~[0, 1]^C$ by solving
\[ \hat{\theta}(D_w) = \mathrm{argmin}_\theta \frac{1}{|D_w|}\sum_{(x_i, y_i) \in D_w} \mathcal{L}(h_\theta(x_i), y_i) \]
where $\mathcal{L}$ is some training loss and $\theta$ represents the parameters of $h$. Different WSOL methods are primarily distinguished by the loss functions and training protocols they use to train $h$.

\noindent
\textbf{(2) Localize.} For each $(x_i, y_i, \textbf{b}_i) \in D_f$, predict a bounding box
\[\hat{\textbf{b}}_i = g(x_i, y_i | h_{\hat{\theta}(D_w)})\] 
according to some procedure $g:~\mathbb{R}^{H \times W \times 3} \times \{1, \ldots, C\} \to \mathbb{R}^4$. Typically $g$ is a simple sequence of image processing operations applied to the feature maps of the trained classifier $h_{\hat{\theta}(D_w)}$. 

\noindent 
\textbf{(3) Evaluate.} 
Let $E$ denote a suitable WSOL error metric which compares the predicted boxes $\{\hat{\textbf{b}}_i\}_{i=1}^{N_f}$ against the ground-truth boxes $\{\textbf{b}_i\}_{i=1}^{N_f}$. Use the validation error $E(D_f| D_w)$ for model selection and hyperparameter tuning and then use a held-out test set $D_\mathrm{test}$ (which is fully labeled like $D_f$) to measure test error $E(D_\mathrm{test}|D_w)$. See \cite{choe2020evaluating} for a discussion of WSOL performance metrics.

\subsubsection{The role of low-shot supervised localization.} 
Without the fully labeled images $D_f$, the WSOL problem becomes ill-posed~\cite{choe2020evaluating}. Since WSOL therefore requires at least a small number of bounding box annotations for validation, it is natural to ask how WSOL compares to few-shot object localization? For our purposes, we define few-shot object localization methods as those which use only $D_f$ for training and validation. Under this definition, the few-shot methods (which use only $D_f$) actually require strictly \emph{less} data than WSOL (which requires both $D_w$ and $D_f$). Since WSOL and few-shot object localization are practical alternatives, it is important to consider them together as in~\cite{choe2020evaluating}. 

\subsection{Label Hierarchies and Label Granularity}
\label{sec:background_granularity}
We define a \emph{label hierarchy} (on a label set $L$) to be a directed rooted tree $H$ whose leaf nodes (\ie nodes $v\in H$ with no children) correspond to the labels in $L$. Edges in $H$ represent ``is-a" relationships, so a directed edge from $u\in H$ to $v \in H$ means that $v$ (\eg ``bird") is a kind of $u$ (\eg ``animal"). We overload $L$ to refer to the label set and to the corresponding set of nodes in $H$. Let $r$ denote the root node of $H$ and let $d(u, v)$ denote the number of edges on the path from $u\in H$ to $v\in H$.

\noindent
\textbf{Coarsening a label.}
Because there is a unique path from the root node $r$ to any leaf node $\ell \in L$, we can ``coarsen" the label $\ell$ in a well-defined way by merging it with its parent node. We define the \emph{coarsening operator} $c_k: H\to H$, which takes any node in the label hierarchy and returns the node which is $k$ edges closer to the root. Thus, $c_0(\ell) = \ell$, $c_1(\ell)$ is the parent of $\ell$, $c_2(\ell)$ is the grandparent of $\ell$, and so on, with $c_{k}(\ell) = r$ for all $k \geq d(r, \ell)$. 

\noindent
\textbf{Coarsening a dataset.} 
We can describe a general ``coarsened" version of $D_w = \{(x_i, y_i)\}_{i=1}^{N_w}$ as $D_w^\mathbf{k} = \{(x_i, c_{k_i}(y_i))\}_{i=1}^{N_w}$ where $\mathbf{k} = (k_1, \ldots, k_{|D_w|})$. If we allow the entries of $\mathbf{k}$ to be chosen completely independently, then we can encounter problems \eg images with multiple valid labels. To prevent these cases, we require $\mathbf{k}$ to be chosen such that $c_{k_i}(y_i)\in H$ is not a descendant of $c_{k_j}(y_j)\in H$ for any $i, j\in \{1, \ldots, N_w\}$. 

\noindent
\textbf{Problem statement.} We can now formalize our key questions: How does $\mathbf{k}$ affect $E(D_\mathrm{test}| D_w^\mathbf{k})$? Are there choices of $\mathbf{k}$ such that $E(D_\mathrm{test}| D_w^\mathbf{k}) < E(D_\mathrm{test} | D_w)$? 

\section{The iNatLoc500 Dataset}\label{sec:new_dataset}

\begin{figure}[t]
\centering 
\includegraphics[trim={0pt 0pt 0pt 0pt},clip,width=\linewidth]{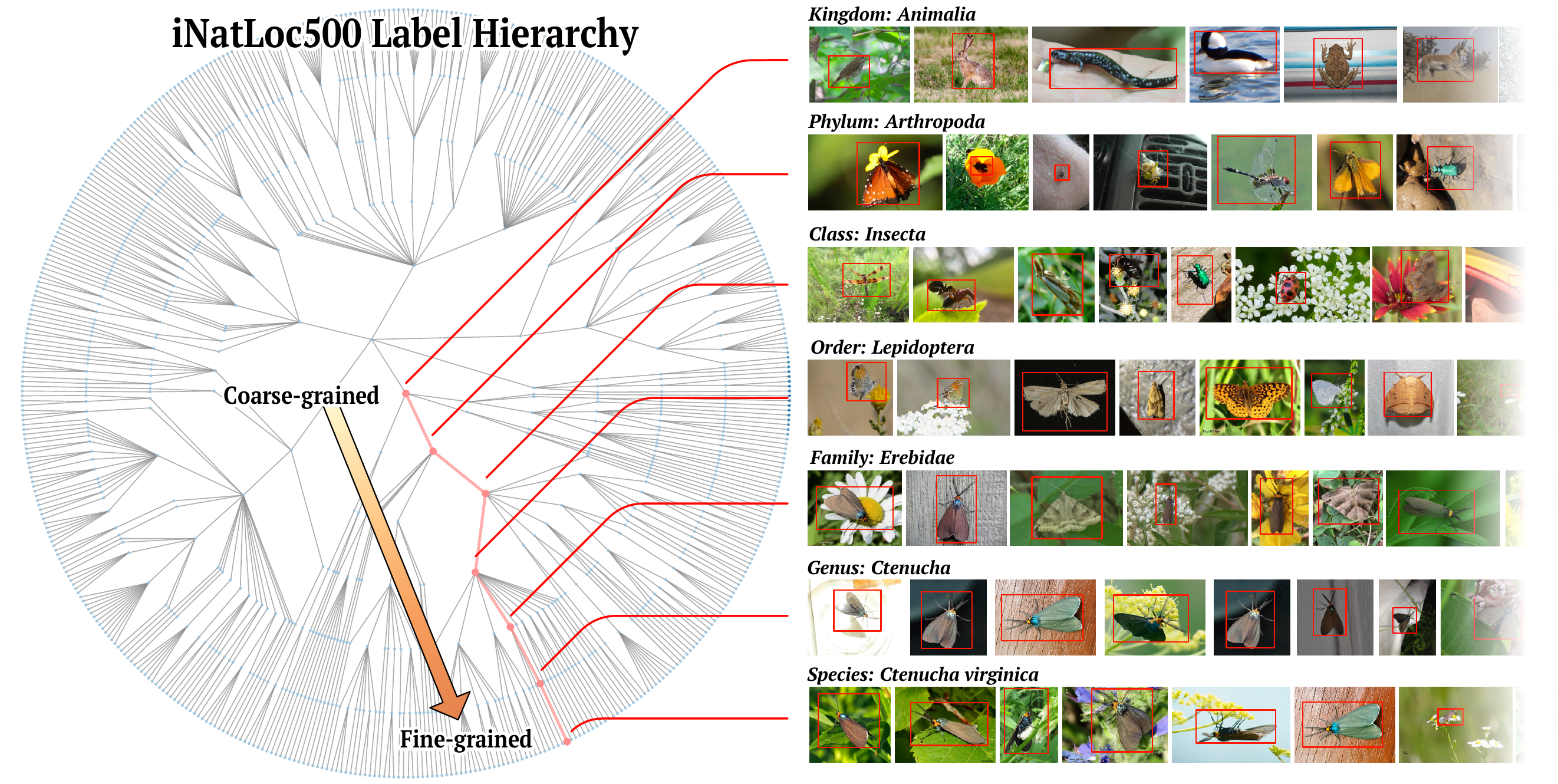}
\caption{Sample images from iNatLoc500 at different levels of the label hierarchy, from coarse (``kingdom") to fine (``species"). Random images from coarse levels of the hierarchy tend to be much more varied than random images ones from finer levels.}
\label{fig:dataset_examples}
\end{figure}

In this section we introduce the iNaturalist Localization 500 (iNatLoc500) dataset, a large-scale fine-grained dataset for weakly supervised object localization. We first detail the process of building the dataset and cleaning the localization annotations. We then discuss the key properties of the dataset and highlight the advantages of iNatLoc500 compared to three WSOL datasets that are currently commonly used (CUB, ImageNet, and OpenImages30k). 

iNatLoc500 has three parts: \texttt{train-weaksup} ($D_w$), \texttt{train-fullsup} ($D_f$), and \texttt{test} ($D_\mathrm{test}$). Each image in the weakly supervised training set ($D_w$) has one image-level category label. Each image in the fully supervised validation set ($D_f$) and test set ($D_\mathrm{test}$) has one image-level category label \emph{and} one bounding box annotation. All bounding boxes have been manually validated. Split statistics are presented in Table \ref{tab:datasets_overview} and sample images from the dataset can be found in Fig.~\ref{fig:dataset_examples}.
The dataset is publicly available.\footnote{\url{https://github.com/visipedia/inat_loc/}}

\subsection{Dataset Construction}\label{sec:dataset_construction}

The iNatLoc500 dataset is derived from two existing datasets: iNat17~\cite{van2018inaturalist} and iNat21~\cite{van2021benchmarking}. Both datasets contain images of plants and animals collected by the citizen science platform iNaturalist~\cite{iNatWeb}. iNat21 is much larger than iNat17 (2.7M images, 10k species vs. 675k images, 5k species), but iNat17 has crowdsourced bounding box annotations. We draw from iNat21 for $D_w$ and we draw from iNat17 for $D_f$ and $D_\mathrm{test}$. 

Full details on the process of constructing iNatLoc500 can be found in the supplementary material, but we note two important design choices here. First, iNat17 did not collect bounding boxes for plant categories because it is often unclear how to draw bounding boxes for plants. Consequently, iNatLoc500 does not contain any plant categories. Second, we set very high quality standards for the bounding boxes. Five computer vision researchers manually reviewed $\sim$ 65k images to ensure the quality of the bounding boxes for $D_f$ and $D_\mathrm{test}$, of which only $51\%$ met our quality standards. Explicit quality criteria and examples of removed images can be found in the supplementary material.

\subsection{Dataset Properties}
\label{sec:dataset_props}
iNatLoc500 is fine-grained, large-scale, and visually diverse. Moreover, iNatLoc500 has a consistent label hierarchy which serves as a reliable proxy for label granularity. We now discuss the importance of each of these properties and contrast iNatLoc500 with existing WSOL datasets.

\noindent
\textbf{Fine-grained categories.} 
Each category in iNatLoc500 corresponds to a different species, and the differences between species can be so subtle as to require expert-level knowledge~\cite{van2018inaturalist}. While there are challenging images in ImageNet and OpenImages30k, most of the categories are coarse-grained \ie relatively few pairs of categories are highly visually similar. For instance, the reptile categories in OpenImages30k (\texttt{lizard}, \texttt{snake}, \texttt{frog}, \texttt{crocodile}) are typically easy to distinguish. In iNatLoc-500 there are 107 reptile species, some of which are highly similar (e.g. \texttt{Chihuahuan spotted whiptail} vs. \texttt{Common spotted whiptail}).

\noindent
\textbf{Consistent label hierarchy.} 
The label hierarchy for iNatLoc500 consists of the following seven tiers, ordered from coarsest to finest: kingdom, phylum, class, order, family, genus, and species. All of the species in iNatLoc500 are animals, so the ``kingdom" tier only has one node (\texttt{Animalia}), which is the root node of the label hierarchy. Every species lies at the same distance from the root. The iNatLoc500 label hierarchy is \emph{consistent} in the sense that all nodes at a given level of the hierarchy correspond to concepts with similar levels of specificity. This means that depth in the label hierarchy measures label granularity. The label hierarchy for CUB is also consistent. However, the taxonomies that underlie ImageNet and OpenImages30k are considerably more arbitrary. For instance, in OpenImages30k some categories are far from the root of the label hierarchy (\eg \texttt{entity/vehicle/land\_vehicle/car/limousine} or \texttt{entity/animal/mammal/carnivore/fox}) while others are close to the root (\eg \texttt{entity/bicycle\_wheel} or \texttt{entity/human\_ear}) despite the fact that there is no obvious difference in concept specificity. 

\noindent
\textbf{Unambiguous label semantics.}
The categories in iNatLoc500 are well-defined in the sense that (for most species) there is little room for debate about what ``counts" as an instance of that species. While the distinctions between species can be quite subtle, each species is a well-defined category. CUB shares this advantage for the most part, but ImageNet and OpenImages30k do not. For instance, OpenImages30k contains the categories \texttt{wine} and \texttt{bottle}. To which category does a bottle of wine belong? (In fact, we find bottles of wine in both categories.) ImageNet is known to have similar issues with ambiguous and overlapping category definitions~\cite{beyer2020we}.

\noindent
\textbf{Visual diversity.}
Like ImageNet and OpenImages30k, iNatLoc500 has a category set which exhibits a high degree of visual diversity. CUB is much more homogeneous, consisting of only birds. Combined with its consistent label hierarchy, the visual diversity of iNatLoc500 enables future work on \eg how localization ability generalizes across categories as a function of taxonomic distance. 

\noindent
\textbf{Large scale.} 
iNatLoc500 is a large-scale dataset, both in terms of the number of categories and the number of training images. CUB and OpenImages30k are considerably smaller on both counts. Large training sets are valuable because they simplify supervised learning. Large training sets also enable research on self-supervised representation learning, which has received little attention thus far in WSOL. We provide a summary of the key dataset statistics in Table~\ref{tab:datasets_overview}. 

\section{Experiments}

In this section we present WSOL results on iNatLoc500 as well as existing benchmark datasets. We also consider few-shot learning baselines based on segmentation and detection architectures. Finally, we use the unique properties of iNatLoc500 to study how label granularity affects localization performance and data efficiency. A summary of the different WSOL datasets can be found in Table~\ref{tab:datasets_overview}.

\subsection{Implementation Details} \label{sec:implementation_details}

\noindent\textbf{Performance metrics.} All WSOL performance numbers in this paper are \texttt{MaxBoxAccV2}, which is defined in \cite{choe2020evaluating}. The only exceptions are the results for OpenImages30k in Table \ref{tab:baseline_results}, which are given in \texttt{PxAP} as defined in \cite{choe2020evaluating}.

\noindent\textbf{Fixed-granularity training.} 
In Sec.~\ref{sec:granularity_experiments} we probe the effect of granularity on WSOL by training on ``coarsened" versions of $D_w$. In the notation of Sec.~\ref{sec:background_granularity}, these can be written $D_w^{k\cdot \mathbf{1}}$ for $k = 1, 2, \ldots$, where $\mathbf{1}$ denotes the ``all ones" vector. This corresponds to merging all leaves with their parent $k$ times. We then run the entire WSOL pipeline from scratch to compute $E(D_\mathrm{test} | D_w^{k\cdot \mathbf{1}})$ for each $k$. To the best of our knowledge this is compatible with all existing WSOL methods. 

\noindent\textbf{Fixed-granularity CAM aggregation.} We also consider a second method for using label hierarchy information to improve WSOL, inspired by \cite{kim2021keep}. Just like traditional CAM, the first step is to train an image classifier using the standard (most fine-grained) label set. However, instead of returning only the CAM for the species labeled in the input image, we return a CAM for each species in the same genus / family / ... / phylum and average them. This ``aggregated" CAM is then evaluated as normal. We abbreviate this method as CAM-Agg.

\noindent\textbf{Hyperparameter search for WSOL methods.} Each time we train a WSOL method we re-tune the learning rate over the set $\{10^{-1}, 10^{-2}, 10^{-3}, 10^{-4}, 10^{-5}\}$ and choose the one that leads to the best \texttt{MaxBoxAccV2} performance on the fully supervised validation set $D_f$. We then report the \texttt{MaxBoxAccV2} performance for the selected model on $D_\mathrm{test}$. We leave all other hyperparameters fixed. Full training details can be found in the supplementary material. 

\noindent\textbf{Non-WSOL methods.} We provide results for the baselines proposed in~\cite{choe2020evaluating} (Center, FSL-Seg), as well as a new few-shot detection baseline (FSL-Det).
``Center" is a naive baseline that simply assumes a centered Gaussian activation map for all images. 
``FSL-Seg" is a supervised baseline that is trained on the $D_f$ split of each dataset. 
The architecture is based on models for saliency mask prediction~\cite{oh2017exploiting}.
Finally, we introduce ``FSL-Det", a few-shot detection baseline for WSOL that is also trained on $D_f$.
It uses Faster-RCNN~\cite{ren2015faster} with the same backbone as other methods (\ie ImageNet-pretrained ResNet-50~\cite{he2016deep}).
Full implementation details can be found in the supplementary material. 
 
\subsection{Baseline Results}
\label{sec:baseline_experiments}

We follow \cite{choe2020evaluating} and evaluate six recent WSOL methods and two non-WSOL methods (Center and FSL-Seg) on iNatLoc500. The results can be found in Table~\ref{tab:baseline_results}. 
We focus our observations on ImageNet, CUB, and iNatLoc500 since OpenImages30k is evaluated using a different task and evaluation metric. 
We first note that our findings on iNatLoc500 reinforce the main results from \cite{choe2020evaluating}, namely that (a) none of the WSOL methods performs substantially better than CAM and (b) FSL-Seg significantly outperforms all WSOL methods. 
Second, if we consider the performance gap between CAM and the Center baseline, we see that simple centered boxes are not as successful on iNatLoc500 (-17.2 \texttt{MaxBoxAccV2}) as they are on CUB (-6.2 \texttt{MaxBoxAccV2}) and ImageNet (-10.3 \texttt{MaxBoxAccV2}). 
This indicates that iNatLoc500 is a more challenging dataset for benchmarking WSOL. 
Finally, we provide results for our few-shot detection baseline (FSL-Det). 
For ImageNet, CUB, and iNatLoc500 we find that FSL-Det is a stronger baseline than FSL-Seg. 
Like FSL-Seg, FSL-Det directly trains on the boxes in $D_f$, whereas the WSOL methods only use those boxes to tune their hyperparameters.  
However, FSL-Det sets a new ceiling for localization performance on these datasets, indicating that current WSOL methods have considerable room for improvement. 

\begin{table}[t]
\caption{
Comparison of WSOL methods. 
Numbers are \texttt{MaxBoxAccV2} for ImageNet, CUB, and iNatLoc500 and \texttt{PxAP} for OpenImages30k.
All results use an ImageNet-pretrained ResNet-50~\cite{he2016deep} backbone with an input resolution of 224x224. WSOL numbers for ImageNet, CUB, and OpenImages30k are the updated results from~\cite{choe2020evaluation_journal}. WSOL numbers for iNatLoc500 are our own, as are the numbers for the baselines (Center, FSL-Seg, FSL-Det). 
FSL baselines use 10 images / class for ImageNet, 5 images / class for CUB, 25 images / class for OpenImages30k, and 25 images / class for iNatLoc500. 
We do not report FSL-Det for OpenImages30k because the evaluation protocol for that dataset requires segmentation masks.
}
\centering
\begin{tabular}{|l|c|c|c|c|} \hline
Method & ImageNet & CUB & OpenImages30k & iNatLoc500 \\ 
\hline
CAM \cite{zhou2016learning} & 63.7 & 63.0 & 58.5 & 60.2 \\
HaS \cite{singh2017hide} & 63.4 & 64.7 & 55.9 & 60.0 \\
ACoL \cite{zhang2018adversarial} & 62.3 & 66.5 & 57.3 & 55.3 \\
SPG \cite{zhang2018self} & 63.3 & 60.4 & 56.7 & 60.7 \\
ADL \cite{choe2019attention} & 63.7 & 58.4 & 55.2 & 58.9 \\
CutMix \cite{yun2019cutmix} & 63.3 & 62.8 & 57.7 & 60.1 \\ \hline
Center & 53.4 & 56.8 & 46.0 & 42.8 \\
FSL-Seg & 68.7 & 89.4 & 75.2 & 78.6 \\
FSL-Det & 70.4 & 95.4 & - & 83.6 \\
\hline
\end{tabular}
\label{tab:baseline_results}
\end{table}

\begin{figure}[t]
\centering
\includegraphics[width=0.49\linewidth]{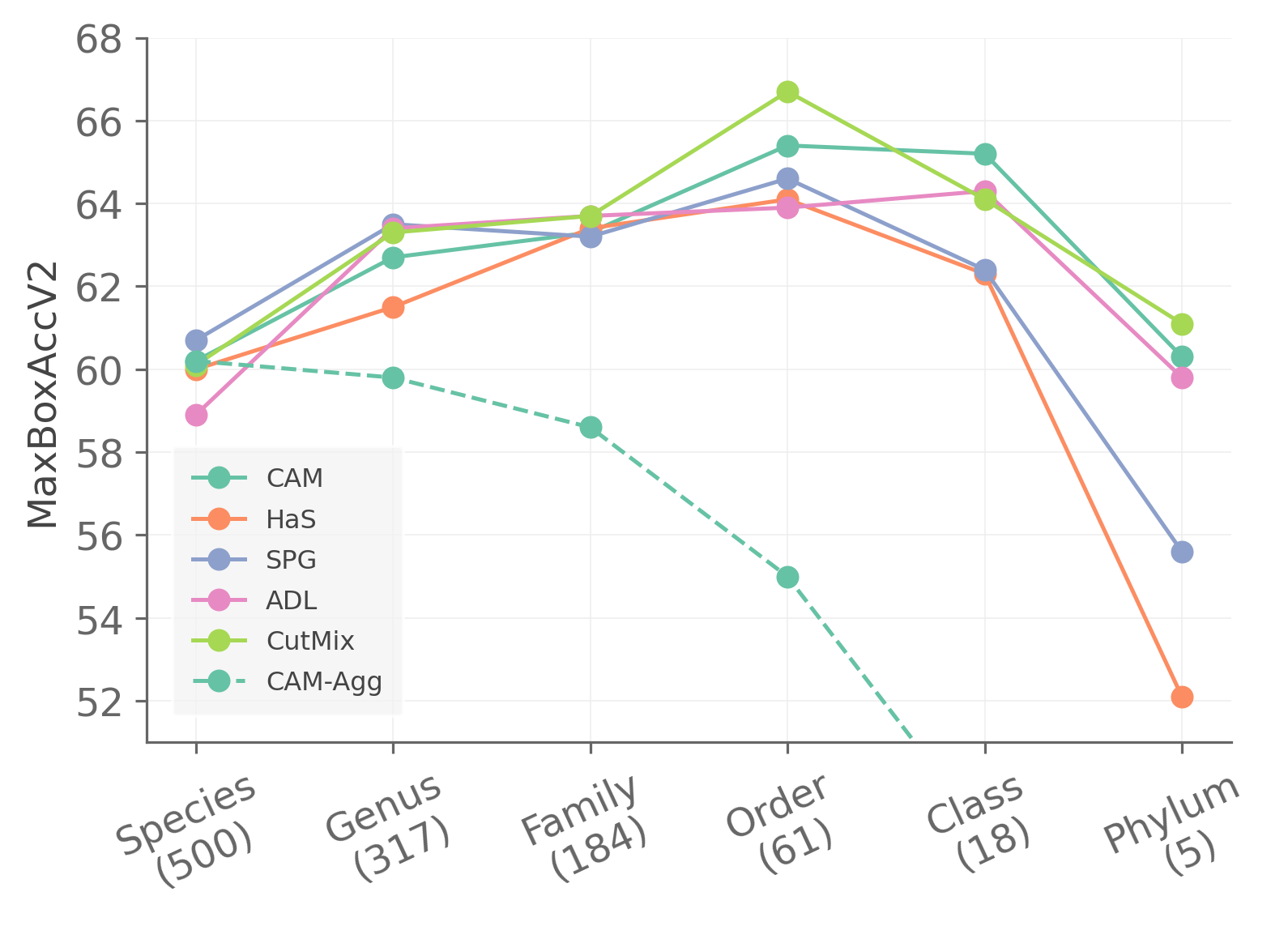}
\includegraphics[width=0.49\linewidth]{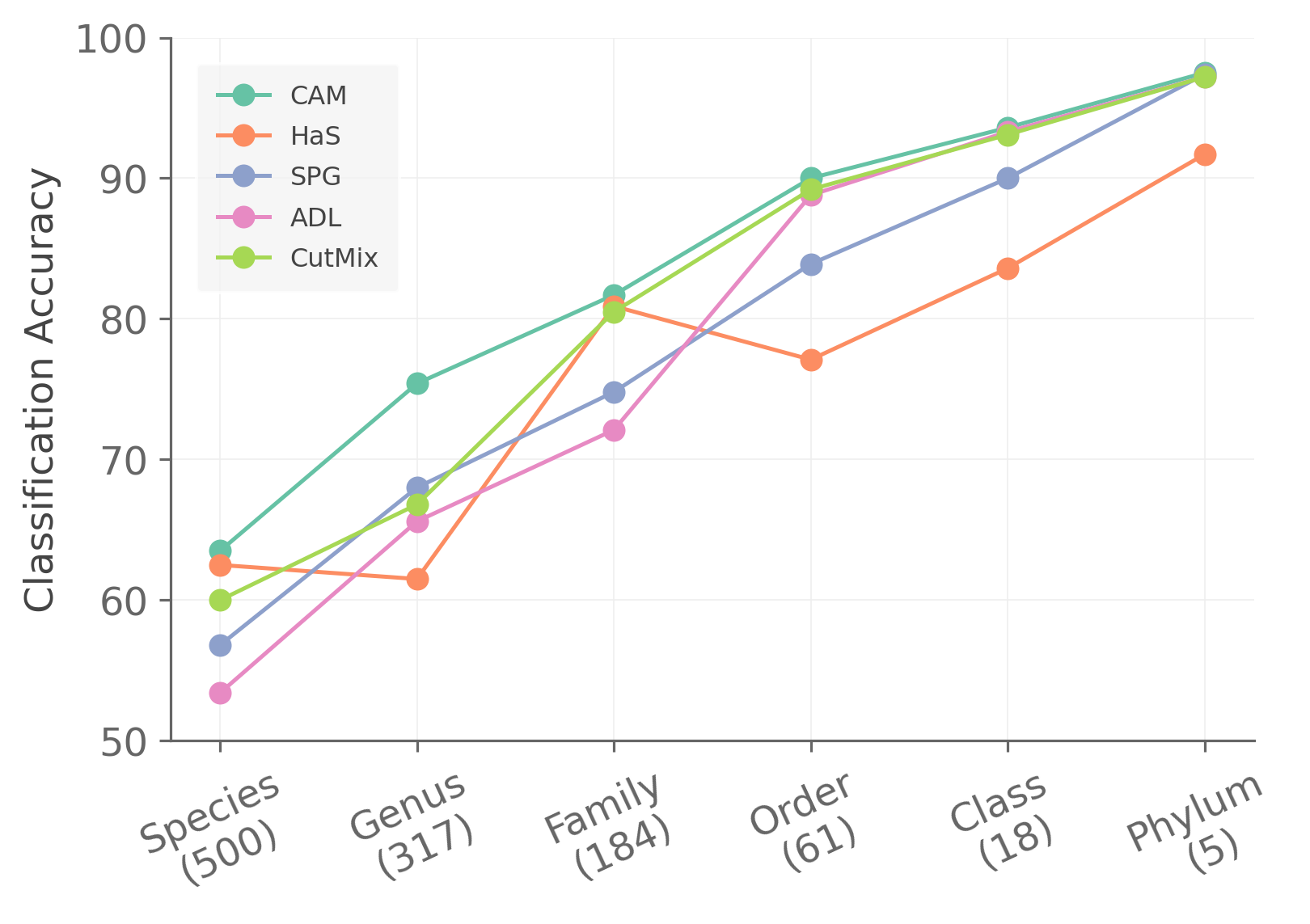}
\caption{
Effect of label granularity of $D_w$ on WSOL performance (left) and classification accuracy (right) for iNatLoc500. The number of categories at each tier is given in parentheses. 
{\bf (Left)} Localization performance suffers when the category labels are either too fine (\eg Species) or too coarse (\eg Phylum). The results on the very left of the plot are the same as those in Table~\ref{tab:baseline_results}. Note that ACoL is excluded due to poor performance -- we suspect it requires more epochs of training than the standard protocol allows for iNatLoc500. We also show results for CAM-Agg (Sec.~\ref{sec:implementation_details}), an alternative method for aggregating hierarchy information in WSOL. 
{\bf (Right)} Each WSOL method trains the image classifier in a different way, but classification accuracy generally increases as the labels become more coarse. Naturally it is easier to distinguish between coarser categories, but it is interesting to note that classification performance is excellent at the phylum level, despite poor localization performance. 
}
\label{fig:granularity_wsol}
\end{figure}

\subsection{Label Granularity and Localization Performance}
\label{sec:granularity_experiments}

iNatLoc500 is equipped with a consistent label hierarchy which allows us to directly study the relationship between label granularity and localization performance. The traditional approach to WSOL on iNatLoc500 would begin by training a classifier on the \emph{species-level} labels, \ie  the finest level in the label hierarchy. However, our hypothesis is that training at the most fine-grained level may not lead to the best localization performance. To study this, we use the fixed-granularity training method discussed in Sec.~\ref{sec:implementation_details}. In particular, we ``re-label" $D_w$ at each level of the label hierarchy using successively coarser categories. We then use each of these re-labeled datasets to train and evaluate different WSOL methods. The results in Fig.~\ref{fig:granularity_wsol}(left) show that coarsening the labels of $D_w$ can significantly boost WSOL performance (e.g. up to +5.1 \texttt{MaxBoxAccV2} for CAM). The numerical values plotted in Fig.~\ref{fig:granularity_wsol}(left) can be found in the supplementary materials. Note that it would be difficult to draw similar conclusions by studying ImageNet or OpenImages30k because their label hierarchies do not measure how fine-grained different categories are -- see Sec.~\ref{sec:dataset_props} for a discussion. Our conceptually simple coarsening approach results in large performance improvements across five different WSOL methods, without any modifications to the model architectures or training losses. 

\noindent 
\textbf{Coarse training beyond iNatLoc500.} 
Fig.~\ref{fig:granularity_wsol}(left) shows that coarse training significantly improves WSOL performance on iNatLoc500. We study the effect of coarse training on FGVC-Aircraft \cite{maji13fine_grained}, CUB \cite{WahCUB_200_2011}, and ImageNet \cite{deng2009imagenet} in the supplementary material. As expected, FGVC-Aircraft and CUB (which have consistent label hierarchies) both benefit from coarse training while ImageNet (which lacks a consistent label hierarchy) does not. 

\noindent
\textbf{Localization performance vs. classification performance.} 
In Fig.~\ref{fig:granularity_wsol}(right) we show the image classification performance for each WSOL method in Fig.~\ref{fig:granularity_wsol}(left) at each granularity level. We see that classification performance and WSOL performance are not necessarily correlated. WSOL performance increases before decreasing at the coarsest level of granularity. Classification performance increases with label coarsening, even at the coarsest level of granularity. 

\noindent
\textbf{An alternative method for incorporating label granularity.}
We also present the performance of CAM-Agg, an alternative method for incorporating granularity information in WSOL (see Sec.~\ref{sec:implementation_details}). 
In our experiments, CAM-Agg underperforms vanilla CAM at every granularity level. As a point of comparison, \cite{kim2021keep} finds that CAM-Agg is better than CAM for CUB but worse than CAM for ImageNet. Our findings suggest that training the model with coarse categories leads to much better localization performance when compared to aggregating the localization outputs for multiple similar fine-grained categories. 

\subsection{Label Granularity and Data Efficiency}

\begin{figure}[t]
\begin{minipage}[c]{0.49\textwidth}
\includegraphics[width=\linewidth]{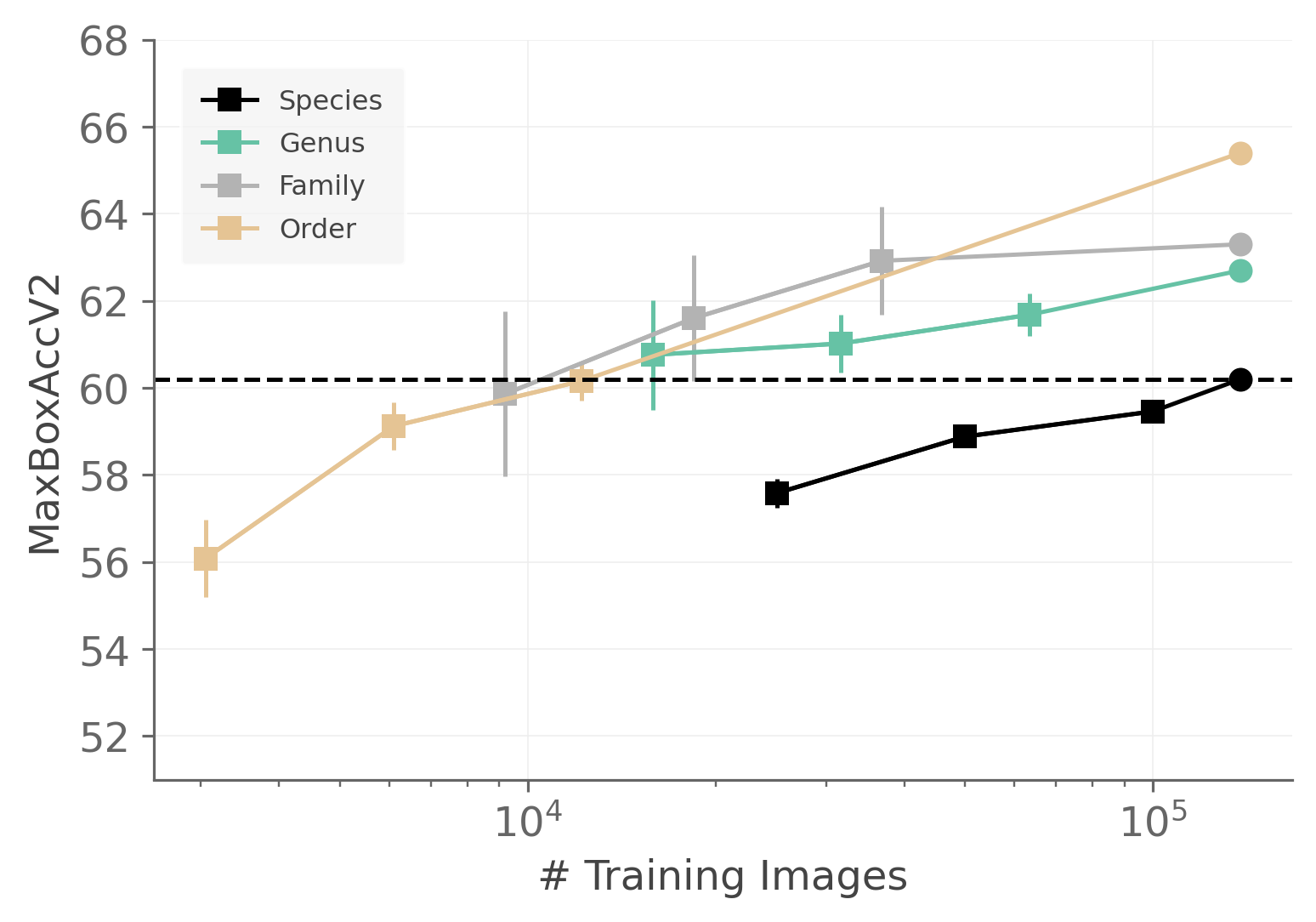}
\end{minipage}\hfill
\begin{minipage}[c]{0.5\textwidth}
\caption{
Effect of the number of training images ($N_w$) on CAM performance for iNatLoc500. 
The dashed line corresponds to the performance of species-level CAM with the entirety of $D_w$. 
Each color corresponds to a different label granularity for $D_w$. Circles at the right of the graph indicate performance using all of $D_w$. 
Squares represent subsampled datasets which use a fixed number of images per category: 50, 100, or 200. All squares have error bars indicating the standard deviation over 5 runs with different randomly sampled subsets of $D_w$.
} \label{fig:inatloc_weaksup}
\end{minipage}
\end{figure}

Most WSOL work makes $D_w$ as large as possible by default, so there has been little attention paid to how the size of $D_w$ trades off against localization performance. In this section we analyze the performance of CAM-based WSOL as a function of the size of $D_w$. We are particularly interested in how label granularity interacts with data efficiency. To study this question, we first pick a granularity level and generate subsampled versions of $D_w$ by choosing, uniformly at random, 50, 100, or 200 images from each category. Note that the size of each subsampled version of $D_w$ depends on the granularity level. For instance, if the categories are the 317 genera, then 50 images per category is $50 \times 317 = 15,850$, compared to $50 \times 61 = 3,050$ images if the categories are the 61 orders.
We present WSOL results for four granularity levels in Fig.~\ref{fig:inatloc_weaksup}. We find that by training at a coarser level, we can obtain better performance with fewer labels. All of the square markers above the dashed line in Fig.~\ref{fig:inatloc_weaksup} correspond to cases where we can achieve better performance than the standard species-level CAM approach using fewer labels. To take one example, by training at the family level we can match the performance of the standard CAM approach by training with 50 images per family (9200 images), a training set reduction of $\sim$15$\times$. 

\section{Discussion}

\begin{figure}[htb]
\centering
\includegraphics[width=1.0\linewidth]{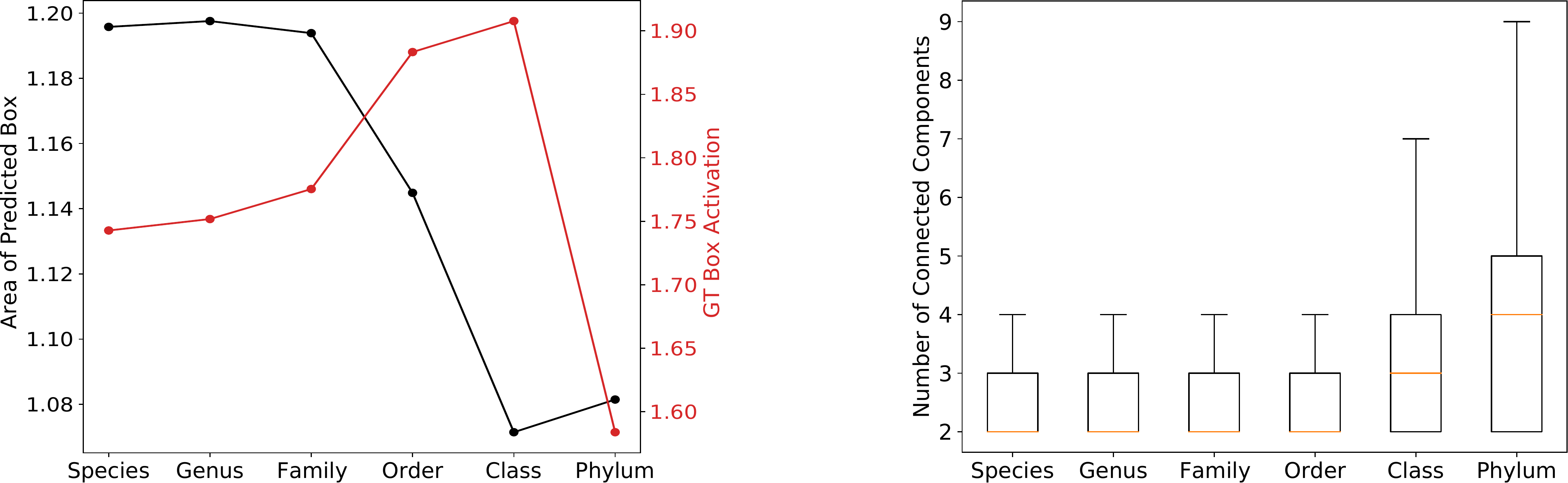}
\caption{
    Analysis of CAM-based WSOL on the $D_f$ split of iNatLoc500. {\bf (Left)} \emph{Black}: Ratio of the area of the predicted box to the area of the ground truth box. \emph{Red}: Ratio of the activation inside the ground truth box to the activation of background pixels. Both curves show medians over the 12.5k images in $D_f$ at each granularity level. {\bf (Right)} Number of connected components in the binarized activation maps at each granularity level. Each box plot shows the distribution over the 12.5k images in $D_f$. See the supplementary material for full details on the construction of these plots. 
}
\label{fig:granularity_qualitative}
\end{figure}

\noindent
\textbf{Why does performance increase as we coarsen the labels?}
In Fig.~\ref{fig:granularity_wsol}(left) we see that five different WSOL algorithms perform better as we coarsen the labels in $D_w$, up until the coarsest level when performance drops. What accounts for this behavior? Our analysis of CAM in Fig.~\ref{fig:granularity_qualitative} provides some clues. Fig.~\ref{fig:granularity_qualitative}(left) shows that the area of the predicted box tends to be larger than the area of the ground truth box, and that their ratio \emph{decreases} towards unity as we coarsen the labels (black curve). 
That is, the predicted box size gets closer to the true box size as we coarsen the labels. 
This casts doubt on a common intuition (which as far as we know has not been empirically investigated before now) that WSOL methods predict smaller boxes for more fine-grained categories \cite{kim2021keep}. 

\noindent
\textbf{Why does performance drop at the coarsest level of granularity?}
In Fig.~\ref{fig:granularity_qualitative}(left) we see that as we coarsen the labels the concentration of \emph{activation} in the ground truth box \emph{increases} before collapsing at the coarsest level (red curve).
Fig.~\ref{fig:granularity_qualitative}(right) shows that the activation maps become highly fragmented at coarser levels. 
Taken together, these two findings suggest that at the coarsest level the activation maps tend to focus more on global image characteristics (\eg land vs. water) than the properties of the foreground object. 
Note that these features are still useful for image classification, as is shown in Fig.~\ref{fig:granularity_wsol}(right). 

\noindent
\textbf{Limitations.} The iNatLoc500 dataset has several limitations. First, it contains only animal categories. These categories are highly diverse, but they are not representative of all visual domains. Second, it is possible that there are errors in the image-level labels provided by the iNaturalist community, though this is expected to be rare as each image has been labeled by multiple people~\cite{van2021benchmarking}. Third, many real fine-grained problems have a long-tailed class distribution but, like other localization datasets, iNatLoc500 is approximately balanced (at the species level). Finally, there is a conceptual limitation in our experiments: the use of a single granularity level across the entire dataset. In fact, it is likely that different images are best treated at different granularity levels. Our work does not address this important topic which we leave for future work. 

\noindent 
\textbf{iNatLoc500 can be used to investigate numerous research agendas beyond traditional WSOL.} For example, $D_w$ was designed to be large enough for self-supervised learning, which has received surprisingly little attention in the WSOL community~\cite{caron2021emerging}. We are also interested in using iNatLoc500 to study whether self-supervised learning methods can be improved by using WSOL methods to select crops~\cite{peng2022crafting}, especially in the context of fine-grained data~\cite{cole2021does}. 
For the object detection community, the clean boxes in iNatLoc500 can (i) serve as a test set for object detectors trained on the noisy iNat17 boxes, (ii) be used to study the problem of learning multi-instance detectors from one box per image, and (iii) be used to analyze the role of label granularity in object detection. Finally, we have seen that hierarchical reasoning can significantly improve localization performance. In the future, we aim to explore methods for automatically determining the most appropriate level of coarseness required for generating representations that best encode object location. 

\section{Conclusion}
We have shown that substantial improvements in WSOL performance can be achieved by modulating the granularity of the training labels, and that coarser-grained training leads to more data-efficient WSOL. We also presented iNatLoc500, a new large-scale fine-grained dataset for WSOL. 
Despite the gains in performance from coarse-level training, iNatLoc500 remains a challenging localization task which we hope will motivate additional progress in WSOL. 

\noindent
\textbf{Acknowledgements.}
We thank the iNaturalist community for sharing images and species annotations. 
This work was supported by the Caltech Resnick Sustainability Institute, an NSF Graduate Research Fellowship (grant number DGE1745301), and the Pioneer Centre for AI (DNRF grant number P1).

%
%
\bibliographystyle{splncs04}
\bibliography{main}

%
%
\newpage
\appendix
\setcounter{table}{0}
\renewcommand{\thetable}{A\arabic{table}}
\setcounter{figure}{0}
\renewcommand{\thefigure}{A\arabic{figure}}
\input{supp_content}

\end{document}

%% file: supp_content.tex
\section{Additional Experiments}

\subsection{Does coarse training help on other datasets?}

In the main paper we show that training on coarser labels significantly improves WSOL performance on iNatLoc500. It is reasonable to ask whether we can replicate this pattern on other datasets. In Fig.~\ref{fig:coarse_training_other} we present results on FGVC-Aircraft \cite{maji13fine_grained}, CUB \cite{WahCUB_200_2011}, and ImageNet \cite{deng2009imagenet}. We give dataset-specific details below.

\subsubsection{FGVC-Aircraft.} The FGVC-Aircraft dataset consists of images of different kinds of aircraft, which are organized into a label hierarchy with the following tiers, ordered from coarsest to finest: manufacturer, family, and variant. Fig.~\ref{fig:coarse_training_other}(top) shows that training with coarser labels improves WSOL performance (+4.3 \texttt{MaxBoxAccV2}). This shows that the benefits of coarse training are not limited to natural world datasets. 

\noindent
\emph{Training details}: We use the best hyperparameters for CUB from \cite{choe2020evaluating}, except that we train for 10 epochs and decay the learning rate every 3 epochs. 

\subsubsection{CUB.}
Fig.~\ref{fig:coarse_training_other}(middle) shows that training with coarser labels improves WSOL performance (+4.4 \texttt{MaxBoxAccV2}). This indicates that our observations on iNatLoc500 in the main paper generalize to images collected under different protocols \ie iNaturalist user photos vs. iconic images crawled from Flickr. Unlike iNatLoc500, we do not see a drop in performance at the coarsest level. This is is consistent with our previous findings because CUB contains only birds, so its hierarchy terminates before reaching the level of granularity where iNatLoc500 performance drops. 

\noindent
\emph{Training details:} We use the filtered version of CUB as described in Sec.~\ref{sec:label_hierarchies} and train with the best hyperparameters for CUB from \cite{choe2020evaluating}. 

\subsubsection{ImageNet.}
\begin{figure}[htb!]
\centering
\includegraphics[width=0.62\linewidth]{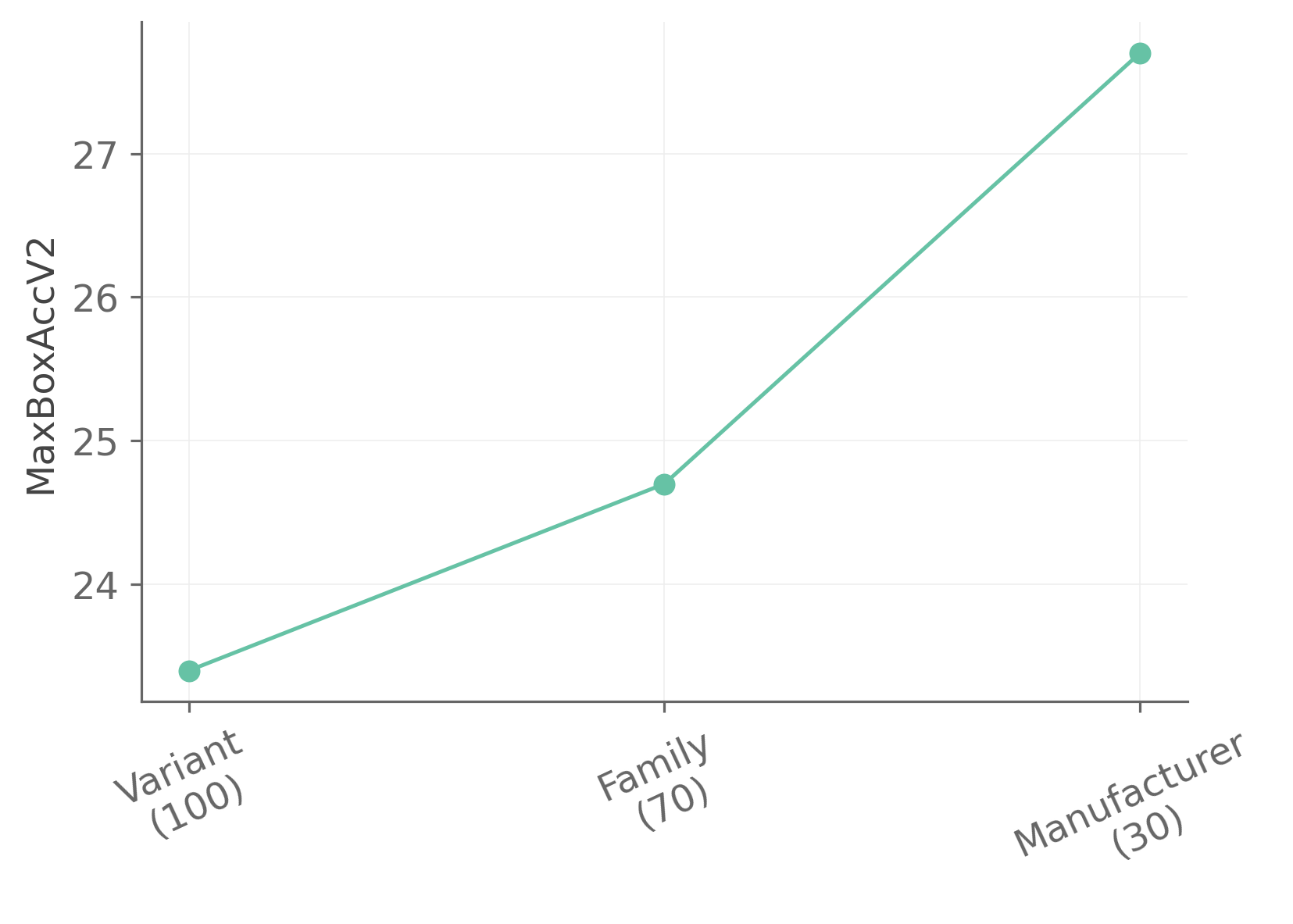}
\includegraphics[width=0.62\linewidth]{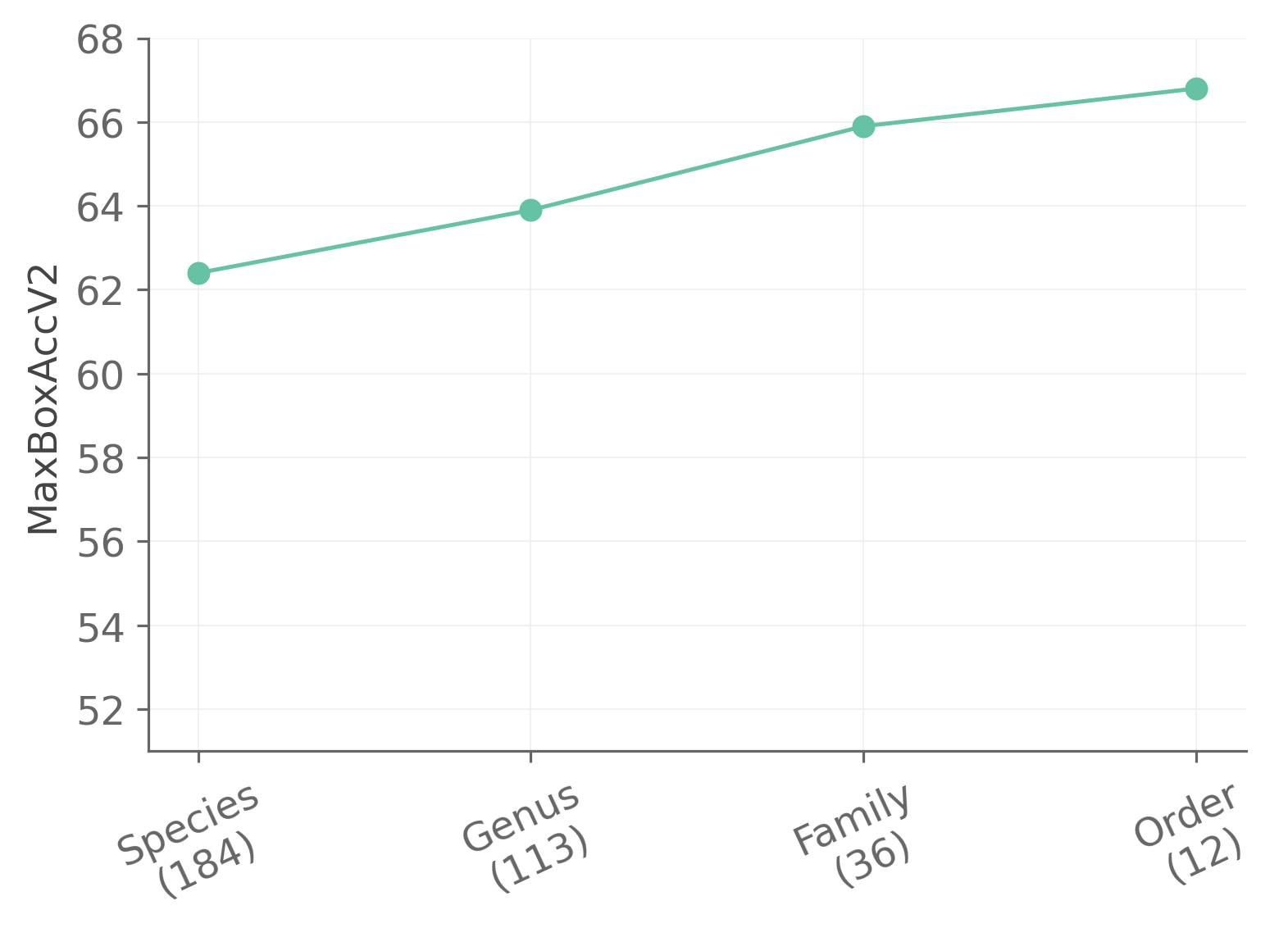}
\includegraphics[width=0.62\linewidth]{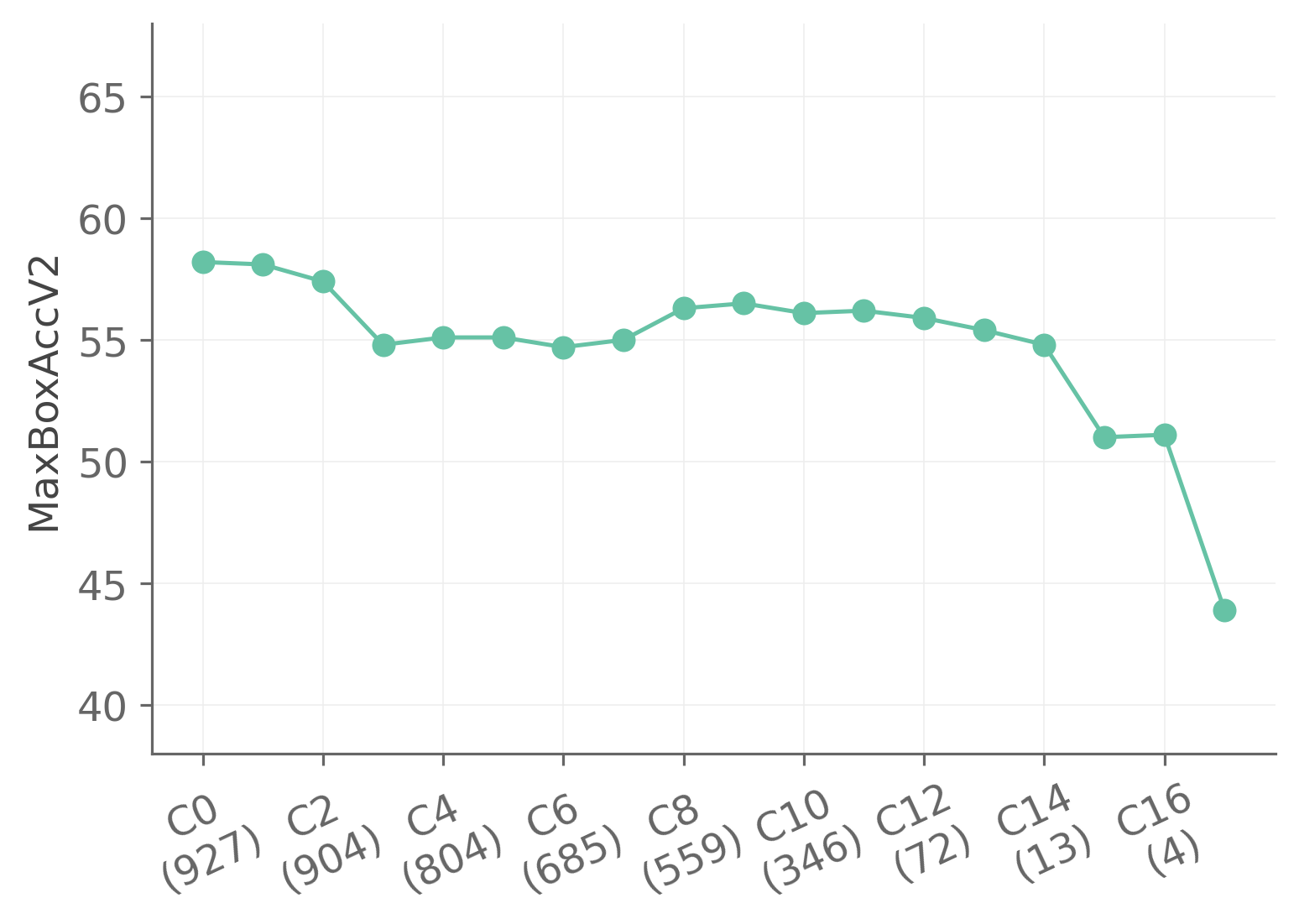}
\caption{
WSOL performance as a function of training set label granularity for FGVC-Aircraft (top), CUB (middle) and ImageNet (bottom). Like iNatLoc500, FGVC-Aircraft and CUB show significant gains at coarser granularities. There is no apparent benefit for ImageNet, which lacks a consistent label hierachy.
}
\label{fig:coarse_training_other}
\end{figure}

 Fig.~\ref{fig:coarse_training_other}(bottom) shows no benefit to coarsening the labels for ImageNet. This is consistent with our claim in the main paper that the ImageNet hierarchy does not measure granularity, and motivates the development of better label hierarchies for datasets like ImageNet in future work. 

\noindent
\emph{Training details:} Unlike iNatLoc500, FGVC-Aircraft, and CUB, the ImageNet label hierarchy has leaf nodes at many different depths. To accommodate this, we must modify the coarsening procedure from the main paper. Instead of coarsening every leaf node at every step, we only coarsen leaf nodes which are at the deepest level of the hierarchy. Each granularity level is named $\texttt{cX}$ where $\texttt{X} \in \{0, 1, 2, \ldots\}$ is the number of times this coarsening has been applied. To speed up training we sample 200 images per category for $D_w$. We use the filtered version of ImageNet as described in Sec.~\ref{sec:label_hierarchies} and train with the best hyperparameters for ImageNet from \cite{choe2020evaluating}.

\subsection{What is the impact of longer training schedules?}

\begin{figure}[htb!]
\centering
\resizebox{0.6\linewidth}{!}{
\includegraphics[width=\linewidth]{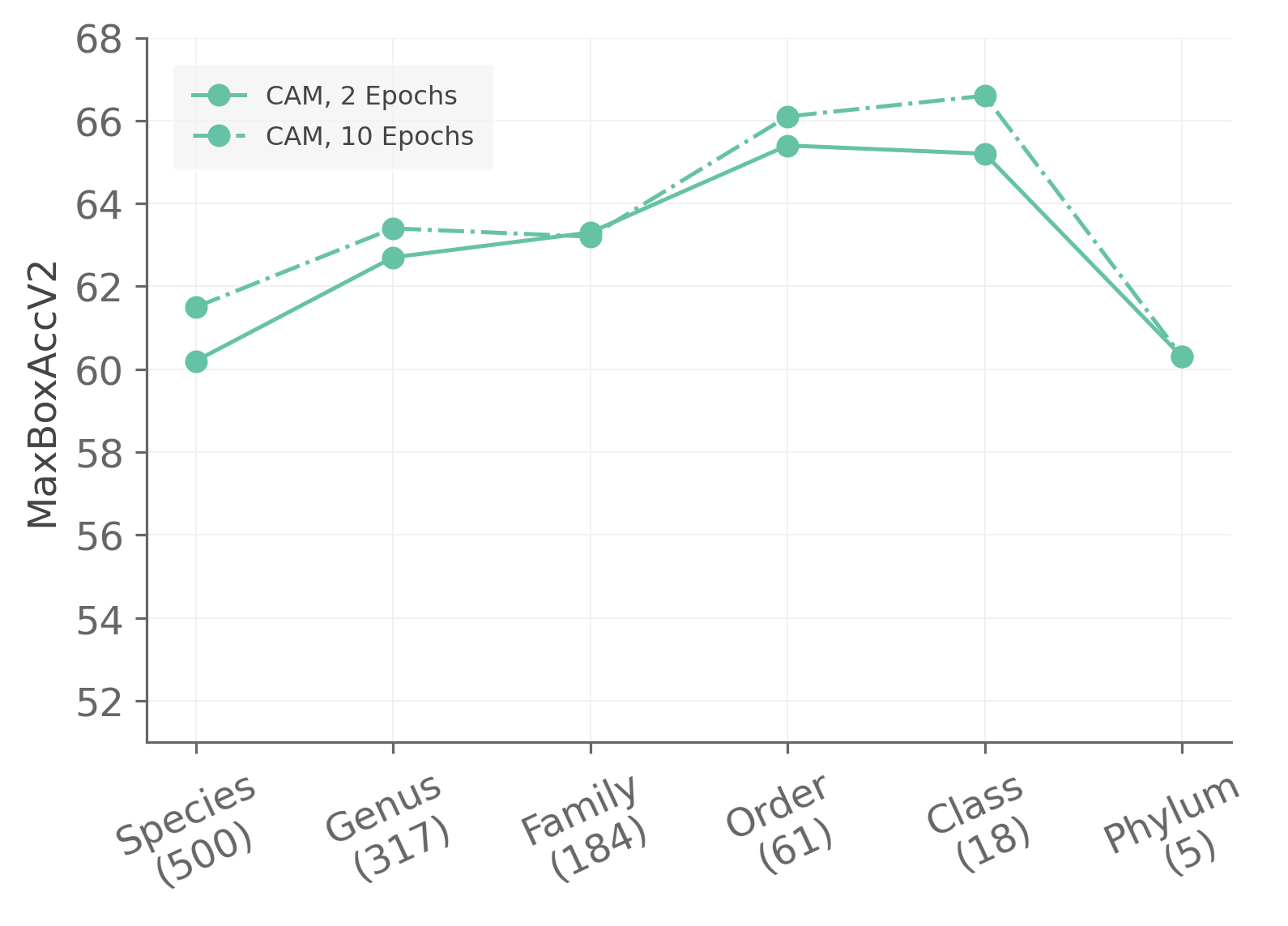}
}
\caption{
Comparison of our standard training schedule (2 epochs, reducing learning rate after 1 epoch) and a longer training schedule (10 epochs, reducing learning rate every 3 epochs) for CAM on iNatLoc500. 
Training for longer does not change the observation that coarser labels result in better localization. 
}
\label{fig:inatloc_granularity_vs_tier_longer_train}
\end{figure}

In \cite{choe2020evaluating}, the authors design their WSOL training schedules so that CUB and OpenImages30k use the same computational budget. They use a budget of 300k images processed, which equates to 50 epochs for CUB and 10 epochs for OpenImages30k. To respect this budget, in the main paper we train on iNatLoc500 for 2 epochs (276k images processed). In Fig.~\ref{fig:inatloc_granularity_vs_tier_longer_train} we see that a longer training schedule can improve performance slightly (Family, Phylum) or significantly (Species, Genus, Order, Class). However, the pattern is the same whether we train for 2 epochs or 10 epochs, \ie performance drops for labels that are too coarse or too fine. 

\subsection{How does WSOL performance depend on the IoU threshold?}

\begin{figure*}[htb!]
\centering 
\includegraphics[width=0.49\linewidth]{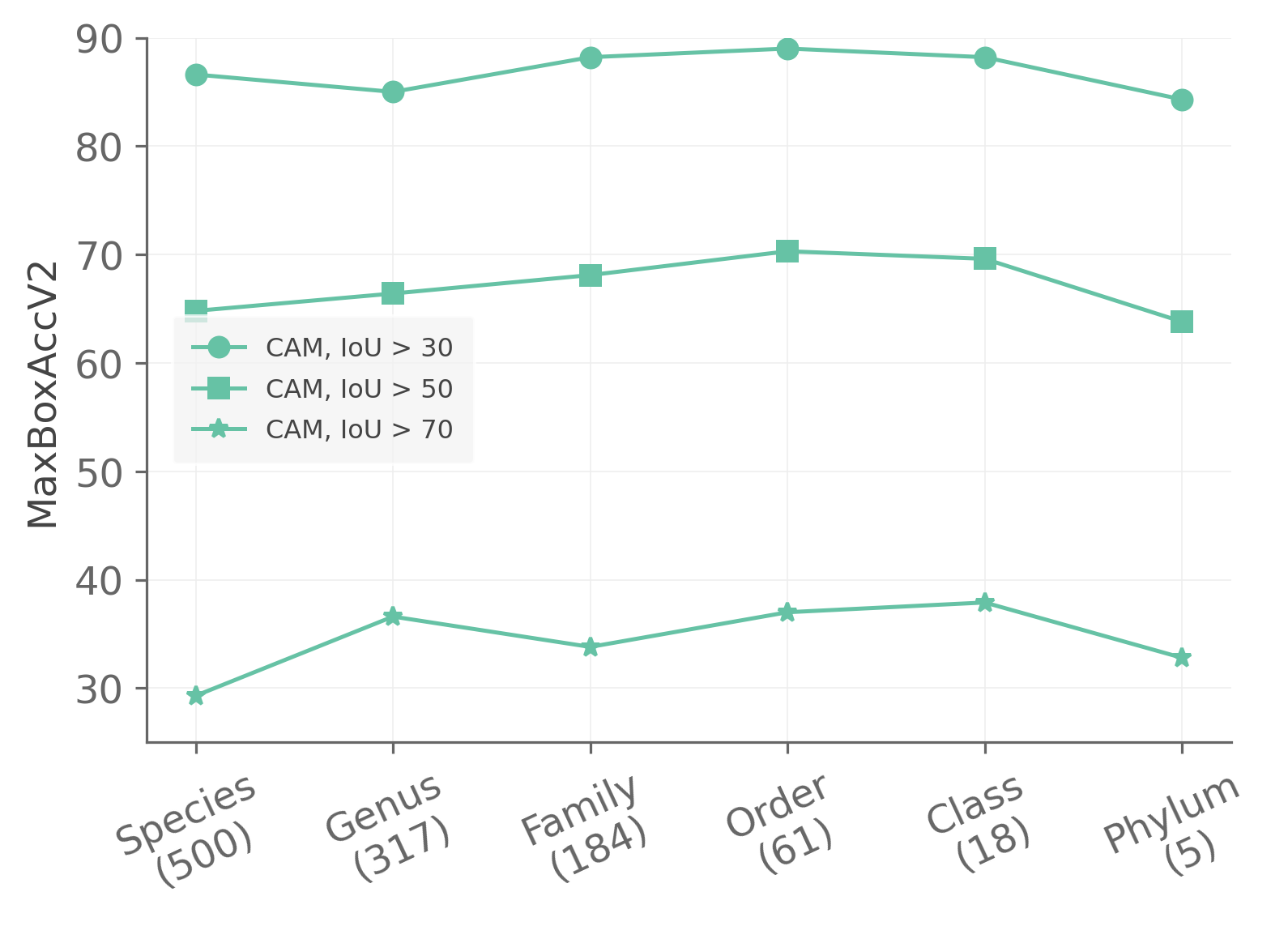}
\includegraphics[width=0.49\linewidth]{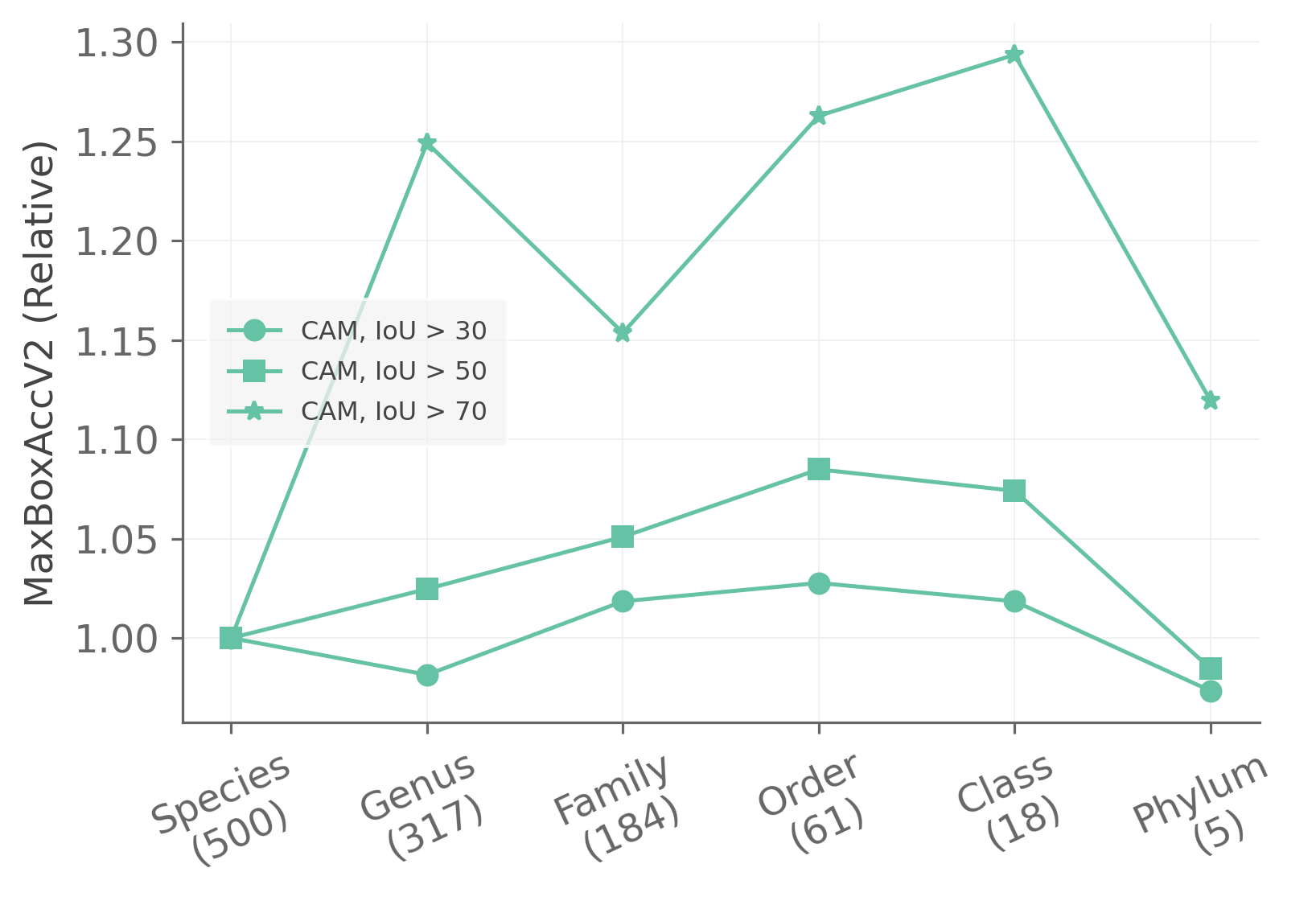}
\caption{
CAM performance on iNatLoc500 as a function of label granularity and IoU threshold. The left panel shows absolute performance. The right panel shows performance relative to the species-level performance, which is the traditional baseline approach. More specifically, the right panel is generated by normalizing each curve in the left panel by its left-most endpoint. 
}
\label{fig:inatloc_granularity_vs_tier_iou_breakdown}
\end{figure*}

Throughout the main paper we use the \texttt{MaxBoxAccV2} metric proposed by \cite{choe2020evaluating}. This metric averages performance over three IoU thresholds: 30\%, 50\%, and 70\%. 
In Fig.~\ref{fig:inatloc_granularity_vs_tier_iou_breakdown} we show the performance of CAM on iNatLoc500 separately for each IoU threshold. 
Not surprisingly, we see that performance decreases significantly as the IoU threshold becomes more demanding (\ie larger). 
We also observe that, regardless of the IoU threshold, the best performance is obtained at a label granularity that is neither too fine nor too coarse. In the right panel of Fig.~\ref{fig:inatloc_granularity_vs_tier_iou_breakdown} we see that the relative performance improvement is larger for more demanding IoU thresholds. This may be because there is less room to improve for ``easier" IoU thresholds.

\subsection{How stable are the CAM results?}

\begin{figure}[htb!]
\centering
\resizebox{0.6\linewidth}{!}{
\includegraphics[width=\linewidth]{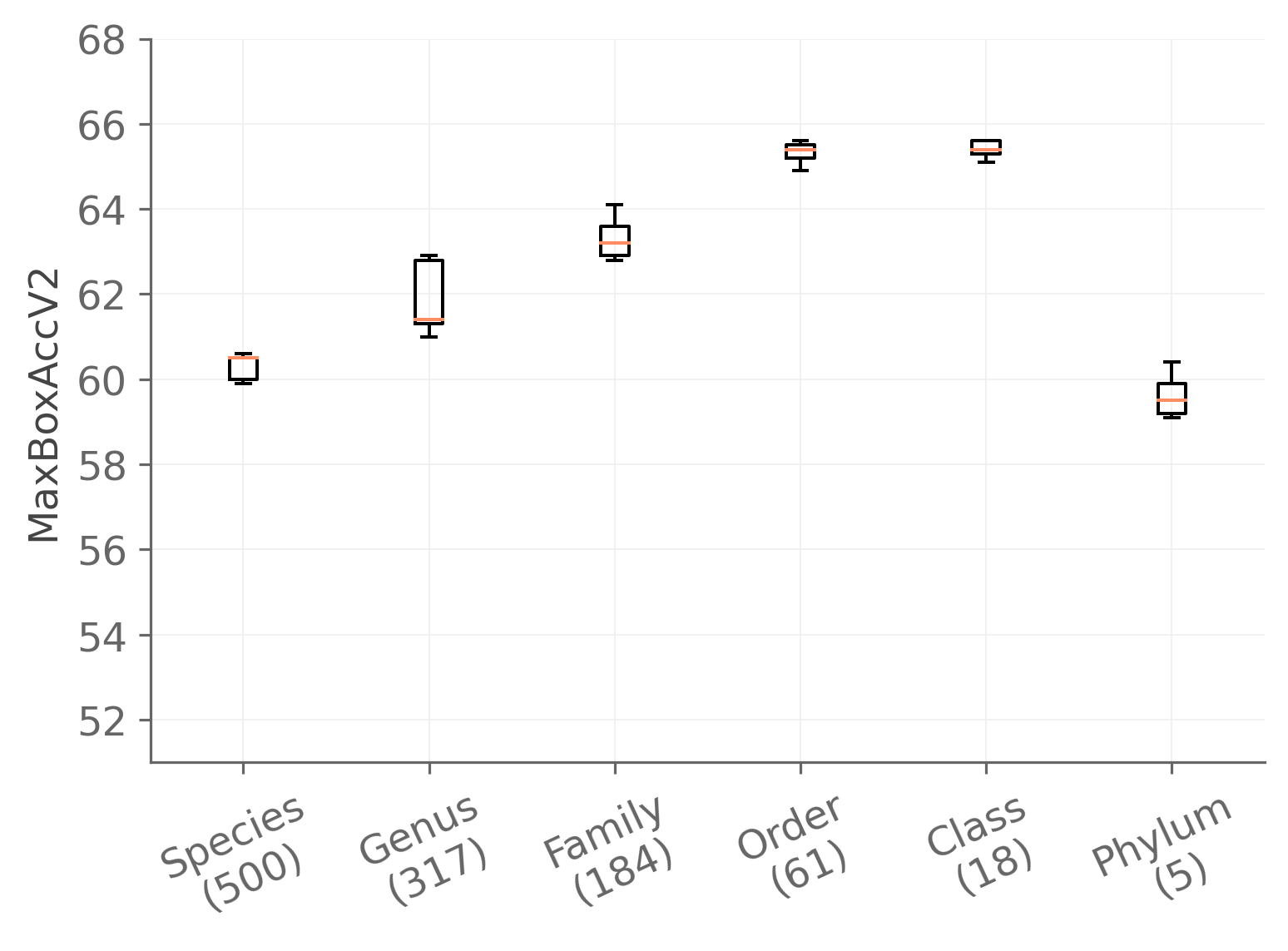}
}
\caption{
Distribution of CAM performance at each granularity level of iNatLoc500 for five runs with identical hyperparameters. 
The orange line denotes the mean. 
}
\label{fig:inatloc_granularity_vs_tier_stability}
\end{figure}

Each result in the main paper is the result of a single run, so it is important to quantify how much test performance varies when we re-train. In Fig.~\ref{fig:inatloc_granularity_vs_tier_stability} we show the results of re-training CAM on iNatLoc500 five times at each granularity level with identical hyperparameters. The standard deviations at different granularity levels range from $\sim 0.2$ to $\sim 0.8$, which is much smaller than the effect sizes we discuss in the main paper. Interestingly, training seems to be most stable for the best-performing coarse-grained levels (order and class), and least stable for the genus level. 

\subsection{What is the effect of additional hyperparameter tuning?}

In their paper, \cite{choe2020evaluating} searches over 30 random hyperparameter sets for each WSOL method. We use a less computationally intensive protocol. For iNatLoc500, we start from their best hyperparameters for ImageNet and re-optimize the learning rate by searching over $\{10^{-5}, 10^{-4}, 10^{-3}, 10^{-2}, 10^{-1}\}$. 
To quantify the performance difference between our reduced hyperparameter search and the full search in \cite{choe2020evaluating}, we conduct both procedures on iNatLoc500 using CAM at the species level. The full hyperparameter search (30 hyperparameter sets) achieves 60.8 \texttt{MaxBoxAccV2} compared to 60.2 \texttt{MaxBoxAccV2} for our abbreviated hyperparameter search (5 hyperparameter sets). As expected, the additional hyperparameter optimization provides an improvement for CAM but the difference is surprisingly modest. We would expect a similar boost to occur for any granularity level. We also note that the gap may be greater for methods with more hyperparameters to tune. We provide the learning rates used in our paper in Table~\ref{tab:learning_rates}. 

\section{Dataset Construction Details for iNatLoc500}\label{sec:dataset_construction_details}

In this section we detail the process of merging and cleaning data from iNat17 and iNat21 to produce iNatLoc500. 

\noindent
\textbf{Species matching.} 
In total there are 4486 species names that occur in both iNat17 and iNat21. We discard any images which do not correspond to a species shared by both datasets. We also omit any species that does not have bounding box annotations. In particular, this means that we discard all plant species, since iNat17 does not have any bounding boxes for plants. 

\noindent
\textbf{Removing duplicate observations.} 
Each image on the iNaturalist platform has an associated \texttt{observation\_id} which corresponds to a unique encounter with an individual plant or animal. We find all observation IDs which occur in both iNat17 and iNat21 and we remove all of the corresponding images from iNat21. It is important to remove duplicates at the \texttt{observation\_id} level instead of the image level, since an iNaturalist observation may be associated with multiple similar but distinct images of the same individual organism. 

\noindent
\textbf{Instance count filtering.} 
Since our focus is object localization (as opposed to detection), any images with multiple bounding box annotations are discarded. 

\noindent
\textbf{Box size filtering.} Any image whose box is smaller than 32 pixels in either dimension is removed. In addition, any image whose box width (height) is more than 96\% of the image width (height) is removed. This step speeds up the annotation process by filtering out a significant number of ``bad" images. Very small boxes are problematic because annotators are more likely to make mistakes, while very large boxes tend to be extreme close-ups. 

\noindent
\textbf{Split considerations.} 
While the majority of observations on iNaturalist are associated with only one image, some do have multiple images. When splitting the fully supervised images into $D_f$ and $D_\mathrm{test}$ we ensure that all of the images for one observation go into exactly one split. This is important because images from the same observation can be highly similar. 

\noindent
\textbf{Manual Annotation.} 
Well-annotated validation and test sets are essential for reliable model selection and benchmarking. The image-level fine-grained class labels reflect the consensus of the iNaturalist community, and like prior iNaturalist datasets \cite{van2018inaturalist,van2021benchmarking} we assume they are correct. However, the bounding box annotations were crowd-sourced with non-expert workers. We therefore manually validate the bounding box annotations for the images in the $D_f$ and $D_\mathrm{test}$ splits. Images with any of the issues listed below were excluded from the dataset. Note that the distribution of images in $D_f$ and $D_\mathrm{test}$ is likely to be somewhat different than the distribution of images in $D_w$ due to this cleaning process. Examples of problematic images are given in the supplementary material.

\noindent
\emph{- Missing instances.} Images with multiple bounding box annotations are filtered out before annotation cleaning.  Unfortunately, the bounding box annotations for an image are sometimes incomplete, which means that an image with one bounding box annotation for a species can contain multiple instances of that species. Images with multiple instances of the labeled species are removed. 

\noindent
\emph{- Inaccurate bounding boxes.} Some bounding boxes are too large or too small, \eg boxes which miss appendages such as legs or tails or boxes which only contain the face of the animal. Images with inaccurate bounding boxes are removed. We also remove any images for which it is unclear whether or not the bounding box is correct, which may occur when an image is blurry or poorly illuminated.  

\noindent
\emph{- Indirect evidence.} iNaturalist accepts images showing \emph{indirect evidence} of an animal (\eg footprints, feathers, droppings), not just images of the animal itself. We omit images which show only indirect evidence of an animal. We also omit images of animal carcasses, which are not uncommon for \eg deer.

\noindent
\emph{- Body part close-ups.} Some images in iNaturalist are clearly intended to show the structure of some specific body part in scientific detail, such as an image of a paw next to a ruler. We omit these images. 

\section{Label Hierarchies}\label{sec:label_hierarchies}

\begin{figure}[htb!]
\centering
\includegraphics[width=0.49\linewidth]{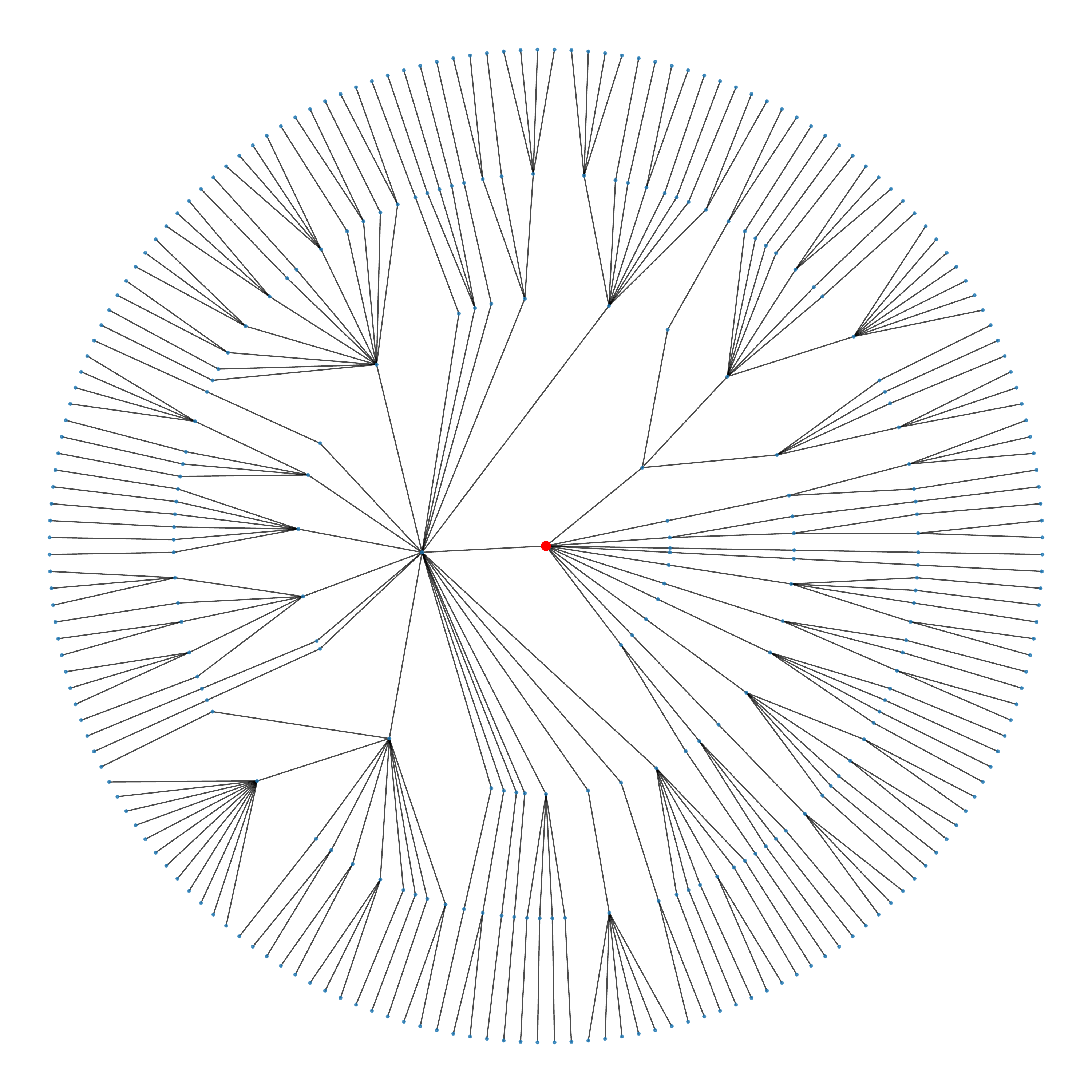} 
\includegraphics[width=0.49\linewidth]{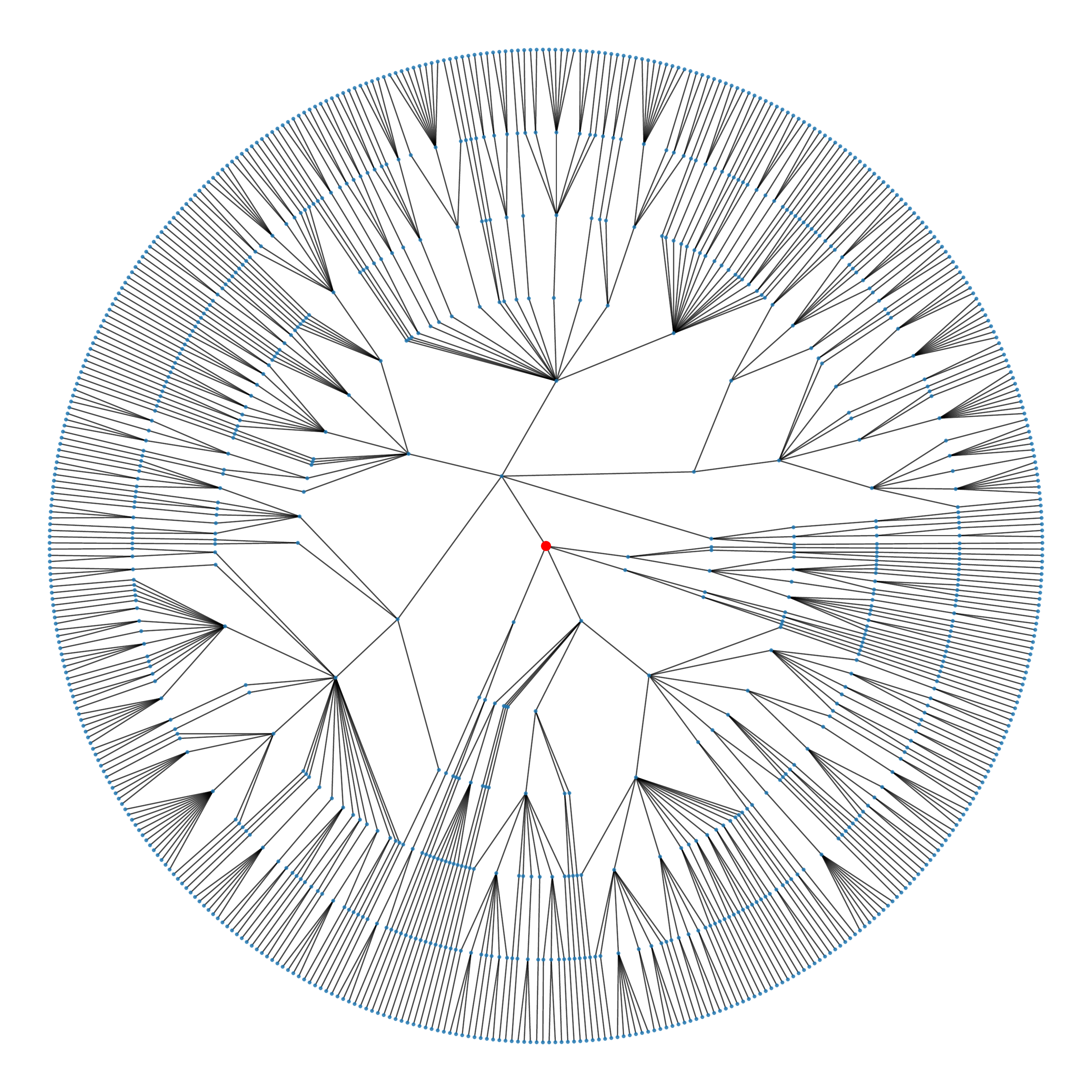} \\
\includegraphics[width=0.65\linewidth]{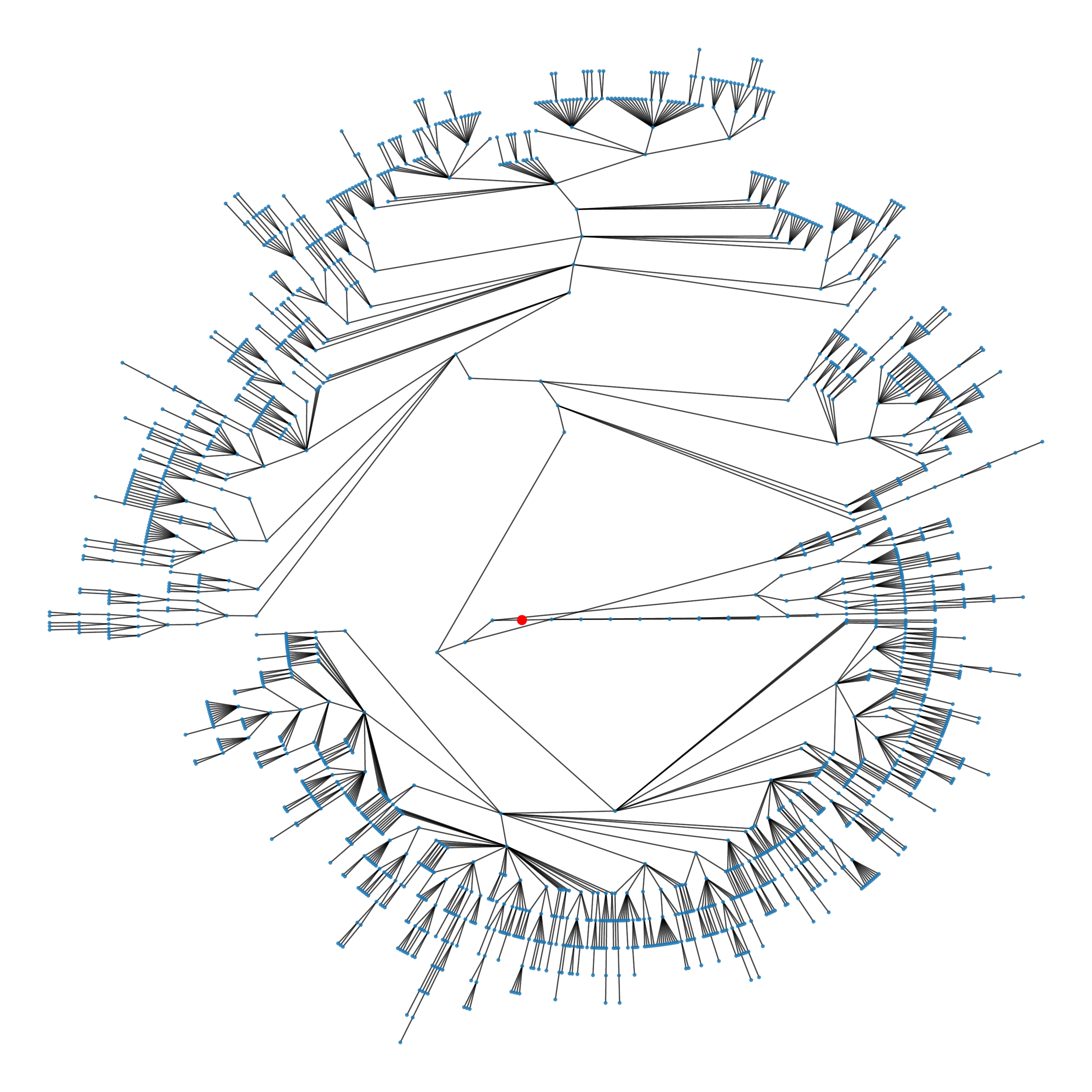} 
\caption{
Label hierarchies for CUB (top left), iNatLoc500 (top right), and ImageNet (bottom). Root nodes are shown in red. The hierarchies for CUB and iNatLoc500 have uniform depths (4 and 6, respectively). The hierarchy for ImageNet is considerably more irregular.
}
\label{fig:tree_viz}
\end{figure}

We visualize the label hierarchies for CUB, ImageNet, and iNatLoc500 in Fig.~\ref{fig:tree_viz}. Producing our final hierarchies for CUB and ImageNet required some care. We give details below.

\subsubsection{CUB.}
CUB was not released with a label hierarchy, so we constructed one. We start by attempting to map each category to a node on the tree of life, like iNatLoc500. CUB consists of 200 bird categories, where some of these categories correspond to species (\eg \texttt{Black-footed Albatross}) and some do not (\eg the genus \texttt{Sayornis} or umbrella terms like \texttt{frigatebird}). We discard any CUB category whose name could not be unambiguously mapped to a single species. By checking these species names against the iNaturalist taxonomy, we obtained the genus, family, order, and class for each species. All bird species belong to the class \texttt{Aves}, so this is the root node of the label hierarchy. Since we retained only species-level categories, every leaf node lies at the same distance from the root node. Our CUB label hierarchy has 184 leaf nodes. 

\subsubsection{ImageNet.}
ImageNet is equipped with a label hierarchy based on WordNet~\cite{miller1995wordnet}. One problem with this hierarchy is that some nodes have multiple parents, which violates the assumptions of the label coarsening procedure outlined in the main paper. We remedy this using a simple greedy approach in which we iterate over the nodes with multiple parents in some fixed (but arbitrary) order and delete all but one parent node. In particular, for each node with multiple parents we perform the following operations: 
\begin{enumerate}
    \item Choose a parent and compute the number of leaf nodes that are still reachable if that parent is retained and the others are deleted. Repeat for each parent.
    \item Keep the parent node for which the greatest number of leaf nodes remain reachable from the root.
    \item Delete the other parent nodes.
    \item Delete any descendants of deleted nodes which are no longer reachable from the root. 
\end{enumerate} 
After executing this process, we obtain a label hierarchy in which each (non-root) node has a unique parent. Our ImageNet label hierarchy has 927 leaf nodes. 

\section{Descriptive Statistics}

\subsection{Class Imbalance and Label Granularity}
\begin{figure}[htb!]
\centering
\resizebox{0.7\linewidth}{!}{
\includegraphics[width=\linewidth]{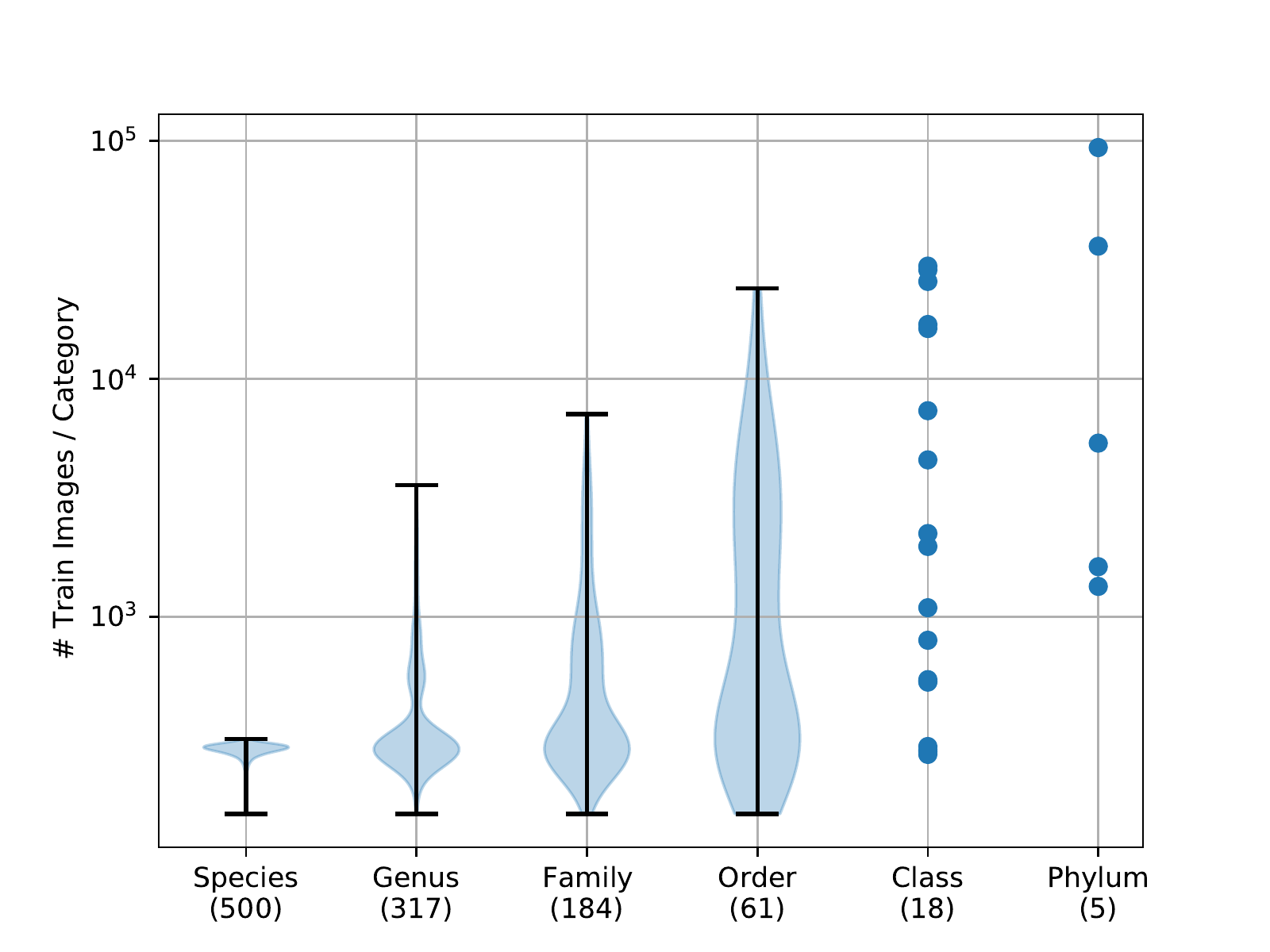}
}
\caption{
Distribution of images over categories at different granularity levels for iNatLoc500. 
We show violin plots for Species, Genus, Family, and Order. 
Class and Phylum contain only a small number of  categories so we can show each point individually. 
See Table~\ref{tab:inatloc_stats_granularity_levels} for summary statistics at each granularity level.
}
\label{fig:inatloc_num_train_per_category_granularity_levels}
\end{figure}

We give basic statistics on the distribution of images over categories for iNatLoc500 at different granularity levels in Table~\ref{tab:inatloc_stats_granularity_levels}. We also visualize the distribution of images over categories at different granularity levels in Fig.~\ref{fig:inatloc_num_train_per_category_granularity_levels}. At the species level, the categories are approximately balanced, but the spread between the largest and the smallest category is much larger for coarser label sets.  

\subsection{Box Size}
\begin{figure}[htb!]
\centering
\includegraphics[width=0.7\linewidth]{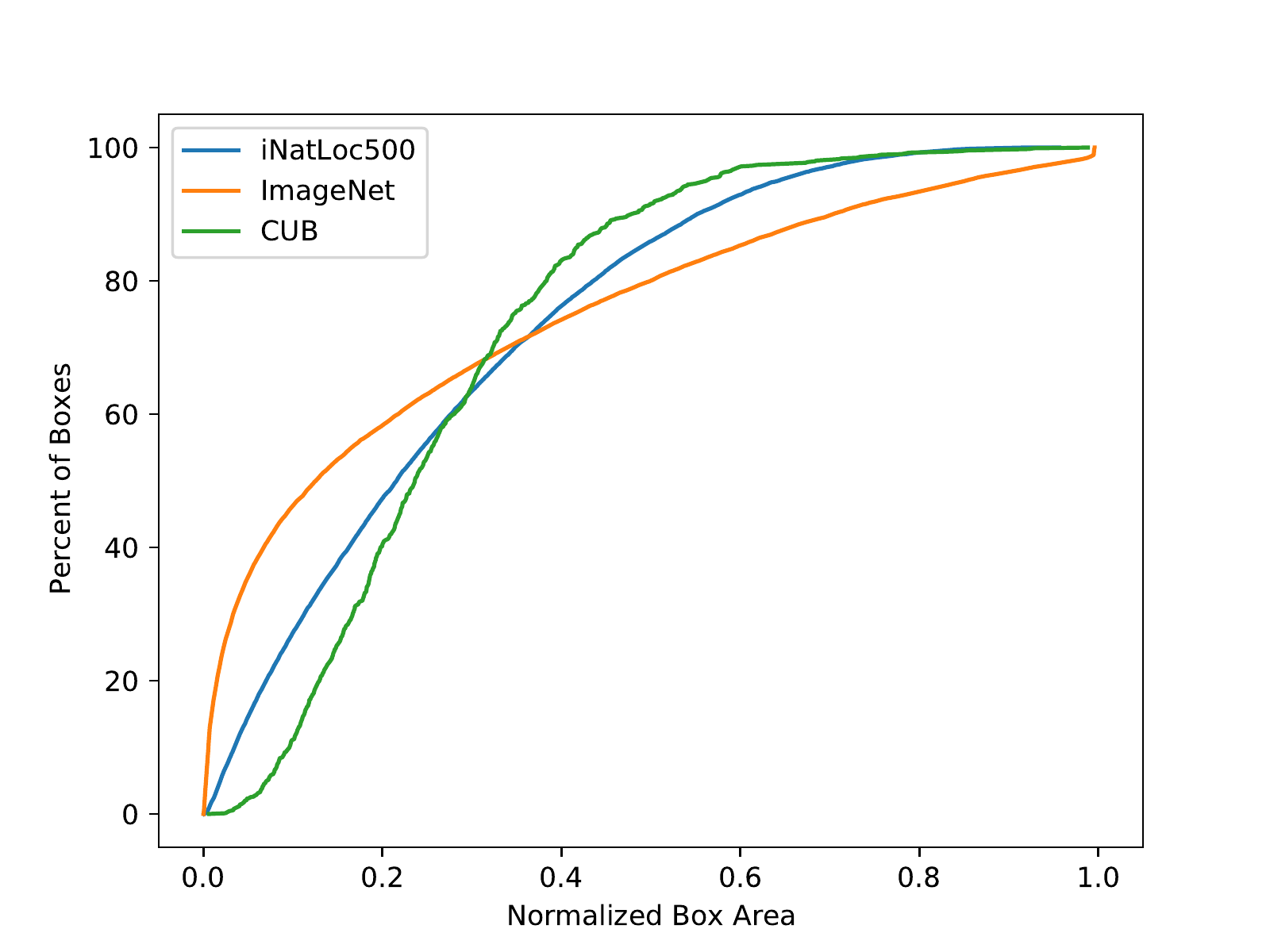}
\caption{
Comparison of CDFs of box sizes for iNatLoc500, ImageNet, and CUB. All box sizes are normalized by the size of the image. 
}
\label{fig:box_sizes}
\end{figure}

In Fig.~\ref{fig:box_sizes} we compare the box size distributions for iNatLoc500, ImageNet, and CUB. For each curve in Fig.~\ref{fig:box_sizes} we compute the area of each box, divide the box areas by the corresponding image sizes, and compute the CDF. The box distribution for iNatLoc500 seems to interpolate between the the box distributions for CUB and ImageNet, \eg iNatLoc500 has more ``small" boxes than CUB but not as many as ImageNet. 

\begin{table}[t]
\caption{
Summary statistics for iNatLoc500 at each granularity level. For each granularity level, we provide the number of categories as well as the minimum, maximum, and mean number of images per category. We also calculate the imbalance factor, which is the size of the largest class divided by the size of the smallest class~\cite{cui2019class}. Refer to Fig.~\ref{fig:inatloc_num_train_per_category_granularity_levels} for a visualization of the distribution of images over categories at each granularity level. 
}
\centering
\footnotesize
\begin{tabular}{|l|c|c|c|c|c|} \hline
Granularity & \# Categories & Min & Max & Mean & Imbalance \\ \hline 
Species & 500 & 149 & 307 & 276 & 2.1 \\
Genus & 317 & 149 & 3575 & 435 & 24.0 \\
Family & 184 & 149 & 7113 & 750 & 47.7 \\
Order & 61 & 149 & 23947 & 2262 & 160.7 \\
Class & 18 & 265 & 29741 & 7666 & 112.2 \\ 
Phylum & 5 & 1345 & 93576 & 27599 & 69.6 \\
\hline
\end{tabular}
\label{tab:inatloc_stats_granularity_levels}
\end{table}

\section{Performance Scores}

For ease of comparison we provide the raw \texttt{MaxBoxAccV2} scores for each WSOL method (and CAM-Agg) at each granularity level in Table~\ref{tab:inatloc_performance_scores}. 

\begin{table}[t]
\caption{
\texttt{MaxBoxAccV2} scores for different WSOL methods trained at different levels of granularity. As in the main paper, ACoL is excluded due to poor performance. We also include provide scores for CAM-Agg, an alternative method of using using granularity information for WSOL described in the main paper. 
}
\centering
\footnotesize
\begin{tabular}{|l|c|c|c|c|c|c|} \hline
Method & Species & Genus & Family & Order & Class & Phylum \\ \hline 
CAM & 60.2 & 62.7 & 63.3 & 65.4 & 65.2 & 60.3 \\
HaS & 60.0 & 61.5 & 63.4 & 64.1 & 62.3 & 52.1 \\
ACoL & - & - & - & - & - & - \\
SPG & 60.7 & 63.5 & 63.2 & 64.6 & 62.4 & 55.6 \\
ADL & 58.9 & 63.4 & 63.7 & 63.9 & 64.3 & 59.8 \\
CutMix & 60.1 & 63.3 & 63.7 & 66.7 & 64.1 & 61.1 \\ \hline
CAM-Agg & 60.2 & 59.8 & 58.6 & 55.0 & 48.8 & 40.1 \\
\hline
\end{tabular}
\label{tab:inatloc_performance_scores}
\end{table}

\section{Implementation Details}

\subsection{Qualitative Analysis of WSOL}
In this section we define the terms used in the qualitative analysis figures in the main paper. In what follows we choose the threshold $t$ to be the optimal threshold for an IoU of 0.50, as defined by \cite{choe2020evaluating}.

\begin{itemize}
    \item \texttt{Area of Predicted Box}: The area of the predicted box divided by the area of the ground truth box. The predicted box is computed using a threshold $t$. 
    \item \texttt{GT Box Activation}: The sum of the heatmap pixels inside the ground truth box divided by the sum of the heatmap pixels outside the ground truth box. 
    \item \texttt{Number of Connected Components}: The number of connected components in the predicted heatmap after it has been binarized with threshold $t$. 
\end{itemize}

\subsection{WSOL Methods}

We consider six standard WSOL methods in this work: CAM~\cite{zhou2016learning}, HaS~\cite{singh2017hide}, ACoL~\cite{zhang2018adversarial}, SPG~\cite{zhang2018self}, ADL~\cite{choe2019attention}, and CutMix~\cite{yun2019cutmix}. We leave the details of those methods to their respective papers. For each WSOL method, we use the training procedures and optimal hyperparameters used by \cite{choe2020evaluating} for ImageNet. The only exceptions are as follows. First, we always use enlarged $28\times 28$ feature maps, instead of letting the choice between 14$\times$14 and $28\times 28$ be an additional hyperparameter. Second, we use a weight decay of $10^{-5}$ instead of $10^{-4}$. Third, we always re-optimize the learning rate by searching over the set $\{10^{-5}, 10^{-4}, 10^{-3}, 10^{-2}, 10^{-1}\}$ and using the value of \texttt{MaxBoxAccV2} on $D_f$ to select the best one. The best learning rate values for each WSOL method and granularity level are provided in Table~\ref{tab:learning_rates} and the method-specific hyperparameters can be found in Table~\ref{tab:hyperparameters}. We summarize the rest of the training details, which match \cite{choe2020evaluating}, below.

\noindent \textbf{Architecture.} 
All methods use an ImageNet-pretrained ResNet-50 backbone with an input resolution of $224\times 224$. 

\noindent \textbf{Image preprocessing.}
Training images are resized to $256 \times 256$, randomly cropped to $224\times 224$, then horizontally flipped with probability 0.5. At test time, images are simply resized to $224\times 224$. All images are normalized according to ImageNet statistics.

\noindent \textbf{Optimization.} 
We train using SGD with Nesterov momentum, a momentum parameter of 0.9, and a batch size of 32. The learning rate for the final linear classifier layer is set to be $10\times$ larger than the learning rate for the rest of the network. For fairness, \cite{choe2020evaluating} trains on each dataset for a number of epochs which equates to processing 300k images. To respect this criterion, we train iNatLoc500 for 2 epochs (276k images processed) and decay the learning rate by a factor of 10 after the first epoch. 

\noindent \textbf{Evaluation.} The search space for the optimal heatmap threshold consists of $1000$ linearly spaced values between $0$ and $1$. Note that all heatmaps are min-maxed normalized before evaluation, so their values fall in $[0, 1]$. Final \texttt{MaxBoxAccV2} numbers are an average over three IoU thresholds: 30, 50, and 70.

\begin{table*}[t]
\caption{
Best learning rates for WSOL methods at different granularity levels. Learning rates are selected from $\{10^{-5}, 10^{-4}, 10^{-3}, 10^{-2}, 10^{-1}\}$ based on the value of \texttt{MaxBoxAccV2} on $\texttt{train-fullsup}$. Note that learning rates for ACoL at coarser granularity levels are omitted due to poor performance. 
}
\centering
\begin{tabular}{|l|c|c|c|c|c|c|} \hline
Method & Species & Genus & Family & Order & Class & Phylum \\ \hline
CAM & $10^{-2}$ & $10^{-1}$ & $10^{-2}$ & $10^{-2}$ & $10^{-2}$ & $10^{-2}$ \\
HaS & $10^{-2}$ & $10^{-1}$ & $10^{-2}$ & $10^{-2}$ & $10^{-2}$ & $10^{-2}$ \\
ACoL & $10^{-3}$ & - & - & - & - & - \\
SPG & $10^{-1}$ & $10^{-1}$ & $10^{-1}$ & $10^{-1}$ & $10^{-1}$ & $10^{-2}$ \\
ADL & $10^{-1}$ & $10^{-1}$ & $10^{-1}$ & $10^{-2}$ & $10^{-2}$ & $10^{-2}$ \\
CutMix & $10^{-2}$ & $10^{-1}$ & $10^{-2}$ & $10^{-2}$ & $10^{-2}$ & $10^{-2}$ \\
\hline 
\end{tabular}
\label{tab:learning_rates}
\end{table*}

\begin{table*}[t]
\caption{
Method-specific hyperparameters used for iNatLoc500. These are the same hyperparameters used by \cite{choe2020evaluating} for ImageNet. 
}
\centering
\begin{tabular}{|l|l|} \hline
Method & Hyperparameters \\ \hline
CAM & N/A \\
HaS & \texttt{drop\_rate} = 0.09, \texttt{drop\_area} = 31\\
ACoL & \texttt{erasing\_threshold} = 0.79\\
SPG & $\delta_l^{B1}$ = 0.02, $\delta_h^{B1}$ = 0.03, $\delta_l^{B2}$ = 0.05, $\delta_h^{B2}$ = 0.47, $\delta_l^{C}$ = 0.29, $\delta_h^{C}$ = 0.36\\
ADL & \texttt{drop\_rate} = 0.68, \texttt{erasing\_threshold} = 0.93\\
CutMix & \texttt{size\_prior} = 0.10, \texttt{mix\_rate} = 0.93\\
\hline 
\end{tabular}
\label{tab:hyperparameters}
\end{table*}

\subsection{Center Baseline}

We perform baseline experiments using the ``center" baseline for WSOL introduced in~\cite{choe2020evaluating}, which simply generates a centered Gaussian heatmap for each image. 
Since \cite{choe2020evaluating} did not fully specify the implementation of their center baseline, our re-implementation may differ slightly. 
We opt for a simple implementation which does not depend on the image shape.
Specifically, we generate an image $C \in \mathbb{R}^{M \times M}$ where 
\[ C_{i,j} = \exp\left(- \frac{((i - \frac{M-1}{2})^2 + (j - \frac{M-1}{2})^2)}{2 \sigma^2}\right)  \]
for the pixel in row $i$ and column $j$.
We then apply min-max normalization to $C$. 
We set $M = 224$ and $\sigma = M/4$. 

Note that in the continuous domain, the value of $\sigma$ would not matter, since, for any $\sigma$, a square centered box of any size could be obtained by choosing the right heatmap threshold. 
In practice, the heatmap threshold is optimized over a fixed grid of values. 
In this case, each value of $\sigma$ yields a different collection of centered boxes, which results in different performance numbers. 

\subsection{FSL-Seg: Few-Shot Localization via Segmentation}

The FSL-Seg baseline for WSOL was introduced in \cite{choe2020evaluating}, but they did not fully specify the implementation details so our approach may differ. Our training protocols are identical to those we use for WSOL methods, except for the modifications described below.

\noindent \textbf{Architecture.} 
Like the WSOL methods, we begin with an ImageNet-pretrained ResNet-50 with an input resolution of $224 \times 224$. We modify the network by replacing the final fully connected layer with a $1\times 1$ convolution layer with a sigmoid activation. Since the feature maps have shape $2048 \times 28 \times 28$, the output of this modified ResNet-50 is a single ``score map" $S \in [0, 1]^{28 \times 28}$. 

\noindent \textbf{Loss.} 
We train using a weighted per-pixel binary cross-entropy loss given by
\begin{align*}
    \sum_{ij} \left[ \frac{Y_{ij}}{\|Y\|_0} \log S_{ij} + \frac{(1 - Y_{ij})}{\|1 - Y\|_0} \log (1 - S_{ij}) \right]
\end{align*}
where $Y \in \{0, 1\}^{28 \times 28}$ is a binary label mask and $\|Y\|_0$ denotes the number of nonzero values in $Y$. This weighting has the effect of equally balancing positive and negative labels. For OpenImages30k, binary label masks are directly available. However, CUB, ImageNet, and iNatLoc500 only have bounding box annotations. For these three datasets we compute $Y$ by converting the bounding box annotations into binary masks. Note that these masks are noisy because most objects do not completely fill their bounding boxes. 

\noindent \textbf{Optimization.}
For each dataset we train for 10 epochs and decay the learning rate by a factor of 10 every 3 epochs. 

\subsection{FSL-Det: Few-Shot Localization via Detection} 

For FSL-Det we use a Faster-RCNN object detection architecture. We use an off-the-shelf TensorFlow Object Detection API training configuration file originally meant for training a Faster-RCNN model on COCO. Other than changing the input image size and the dataset, we do not modify the architecture or any training procedures. The configuration can be found here:

\begin{center}
    {\scriptsize \url{https://github.com/tensorflow/models/blob/65407126c5adc216d606d360429fe12ed3c3f187/research/object\_detection/configs/tf2/faster\_rcnn\_resnet50\_v1\_640x640\_coco17\_tpu-8.config}}
\end{center}

\section{Manual Annotation}

\begin{figure*}[htb!]
\centering 
\includegraphics[width=0.24\linewidth]{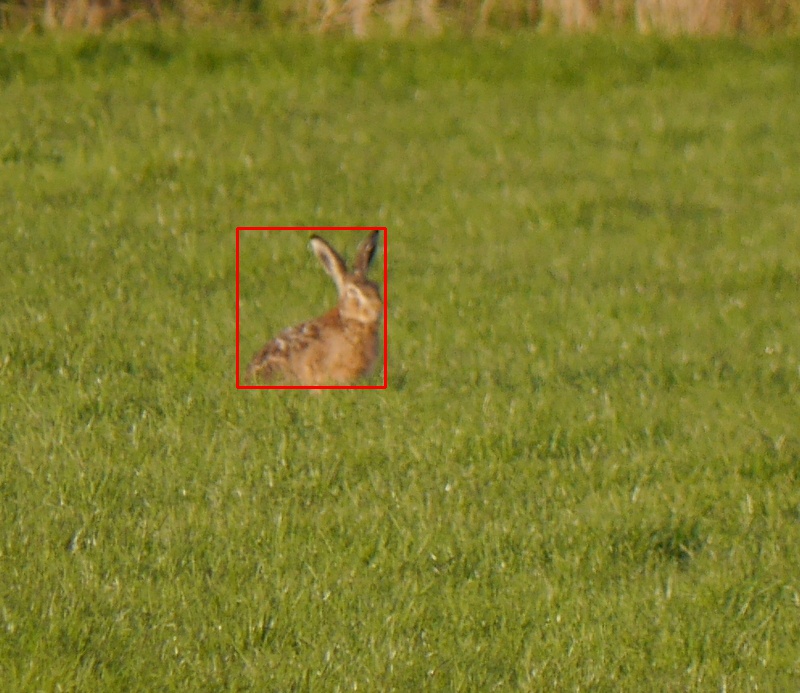}
\includegraphics[width=0.24\linewidth]{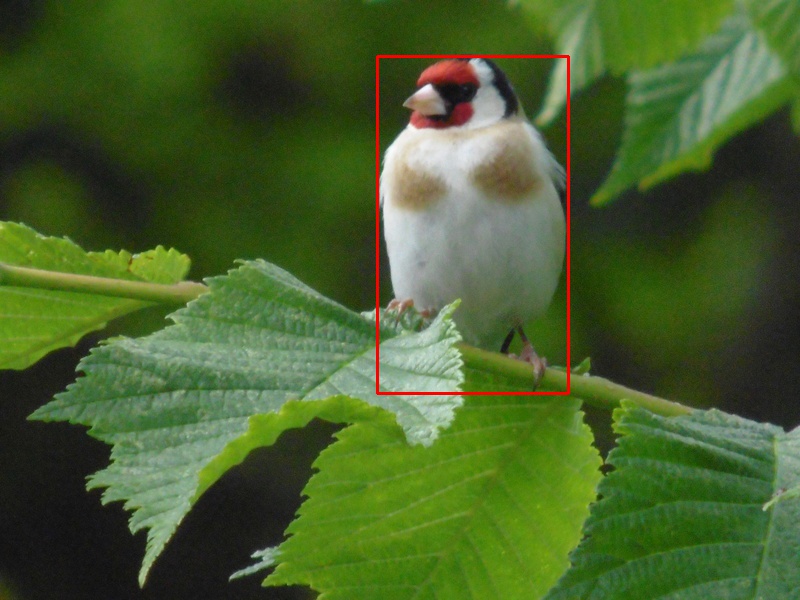}
\includegraphics[width=0.24\linewidth]{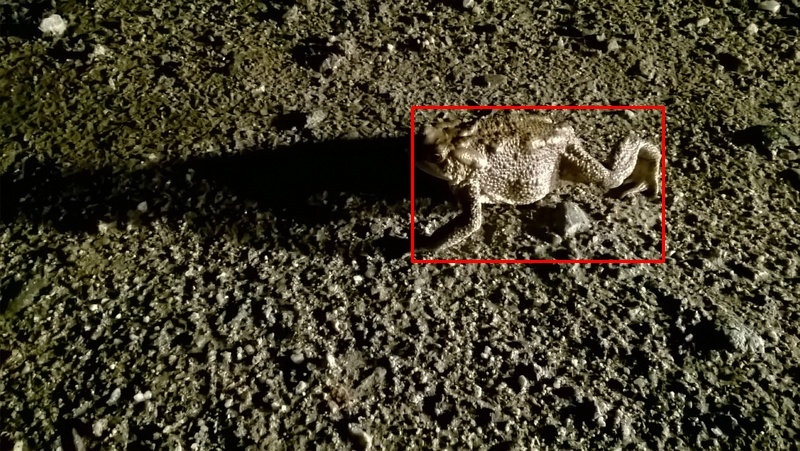} 
\includegraphics[width=0.24\linewidth]{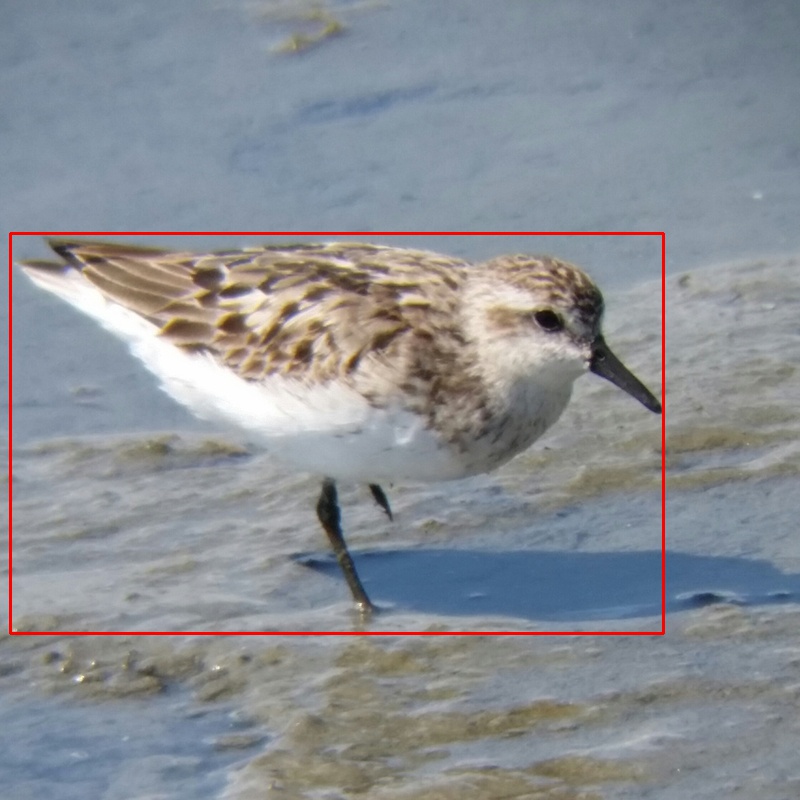}
\\
\includegraphics[width=0.24\linewidth]{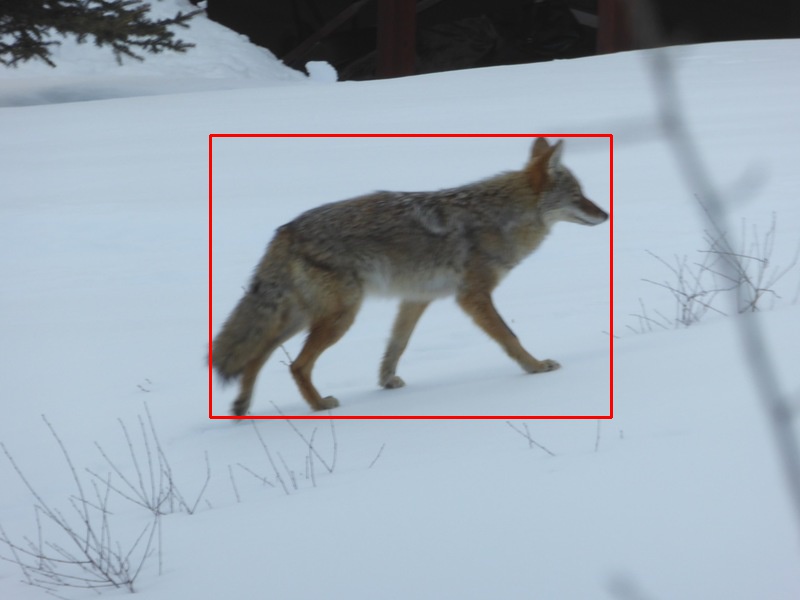}
\includegraphics[width=0.24\linewidth]{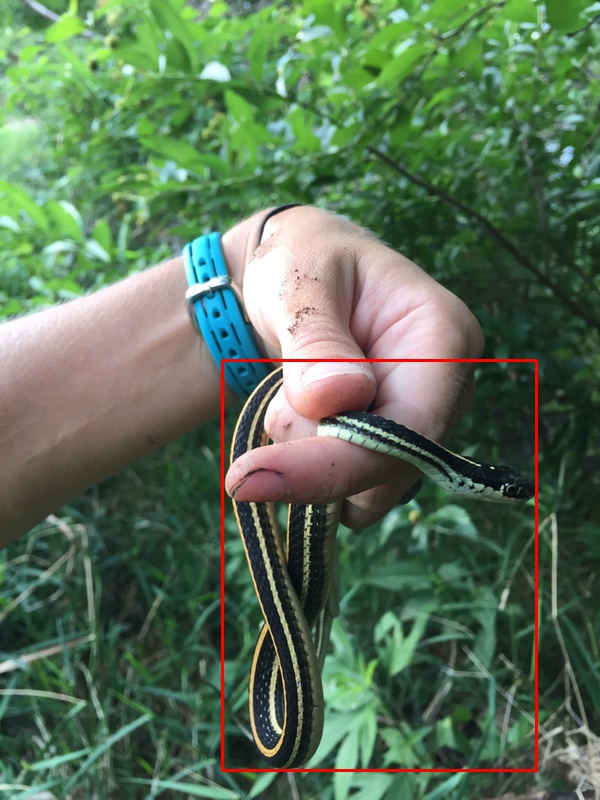}
\includegraphics[width=0.24\linewidth]{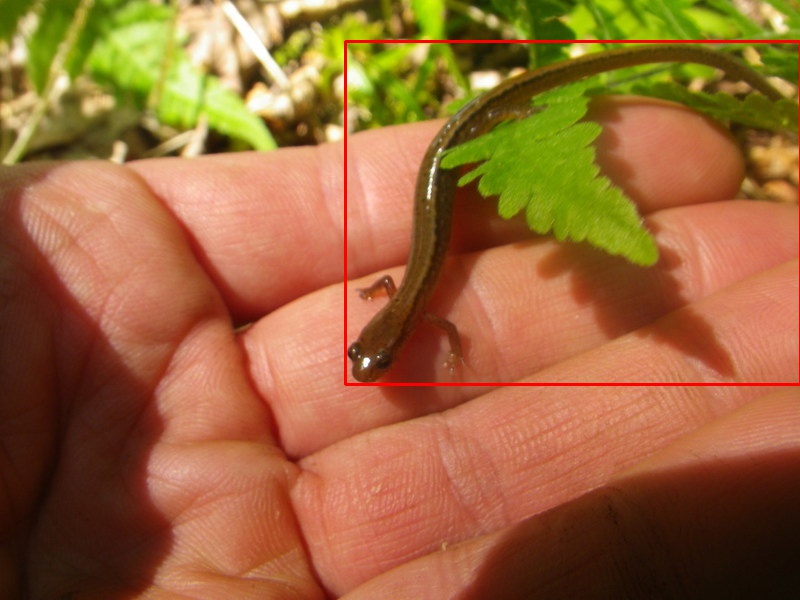}
\includegraphics[width=0.24\linewidth]{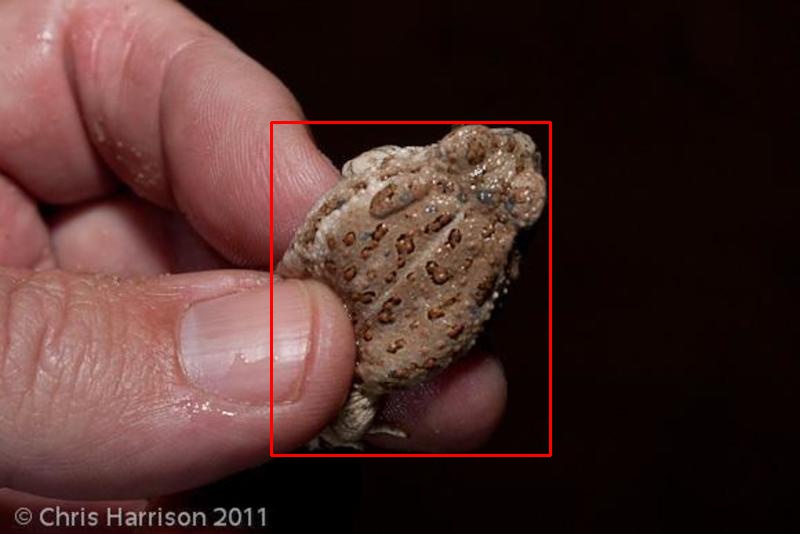}
\\
\includegraphics[width=0.24\linewidth]{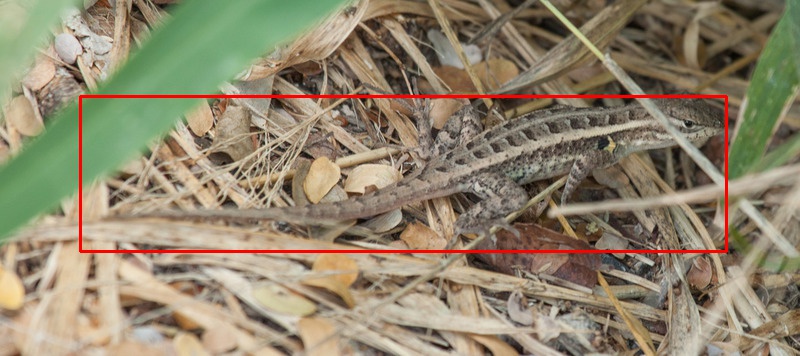}
\includegraphics[width=0.24\linewidth]{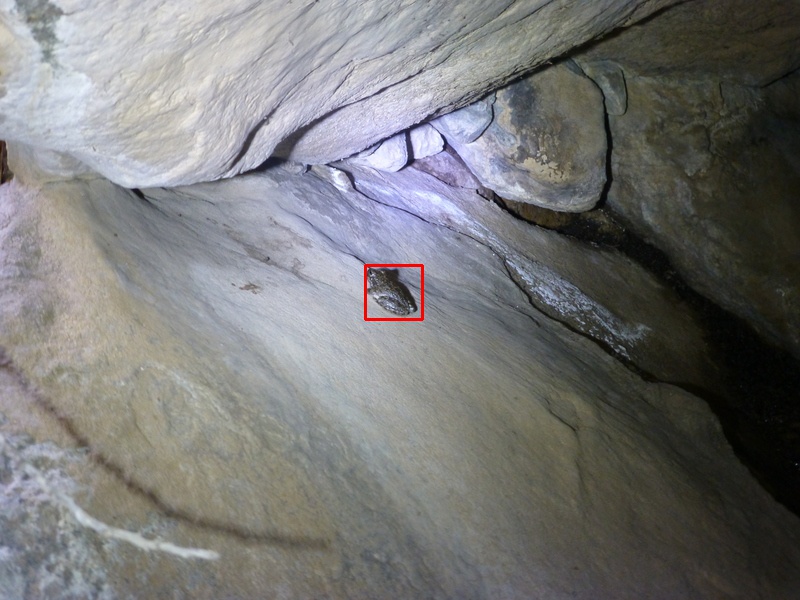}
\includegraphics[width=0.24\linewidth]{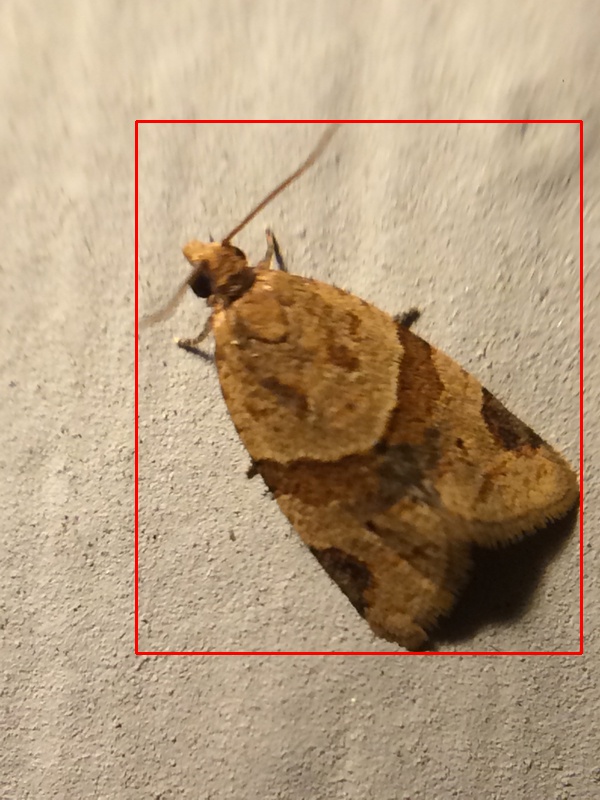}
\includegraphics[width=0.24\linewidth]{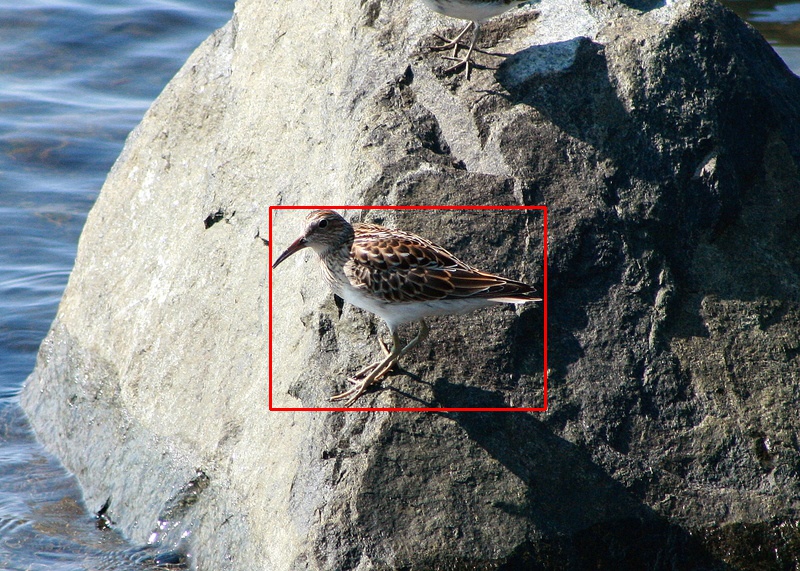}
\\
\includegraphics[width=0.24\linewidth]{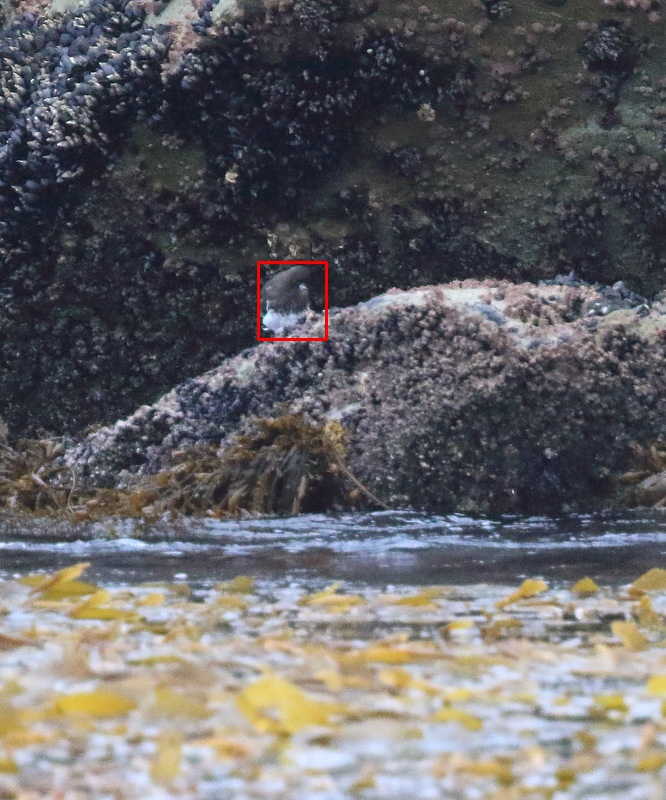}
\includegraphics[width=0.24\linewidth]{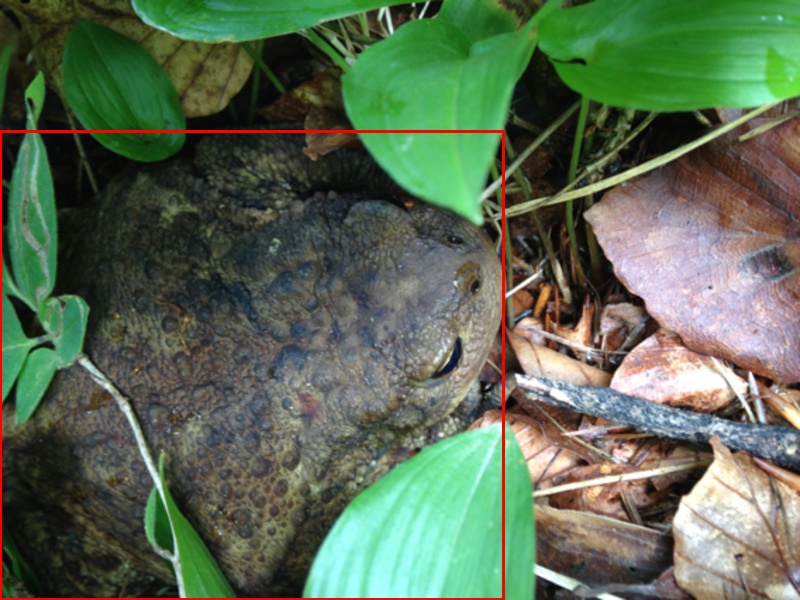}
\includegraphics[width=0.24\linewidth]{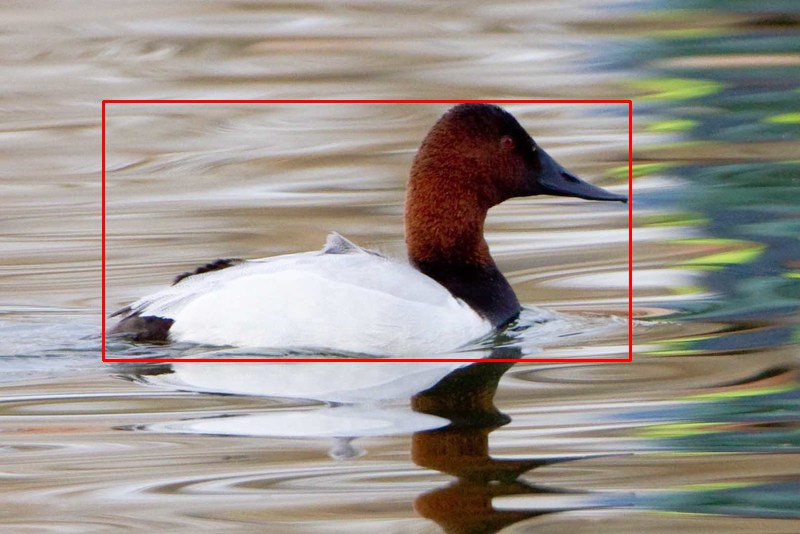}
\includegraphics[width=0.24\linewidth]{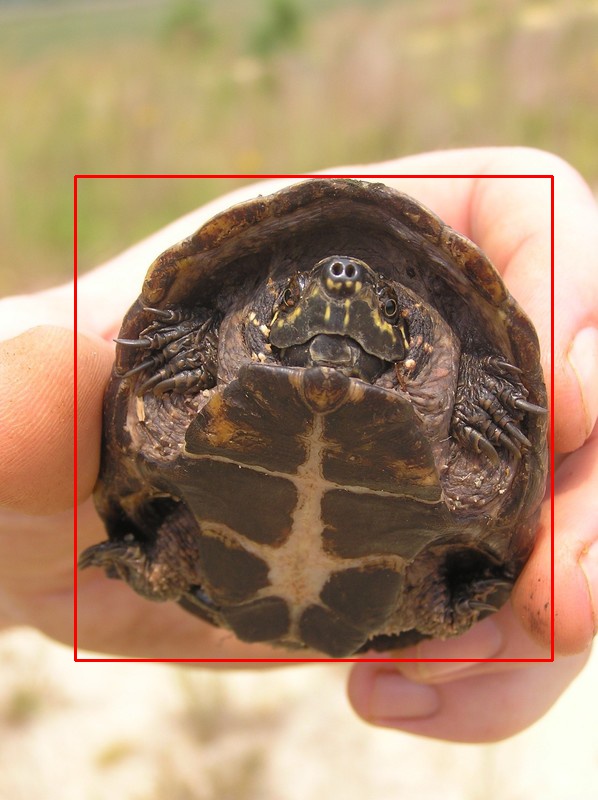}
\caption{
Randomly selected sample images from the iNatLoc500 dataset.
}
\label{fig:rand_examples}
\end{figure*}

\begin{figure*}[htb!]
\centering 
\includegraphics[width=0.24\linewidth]{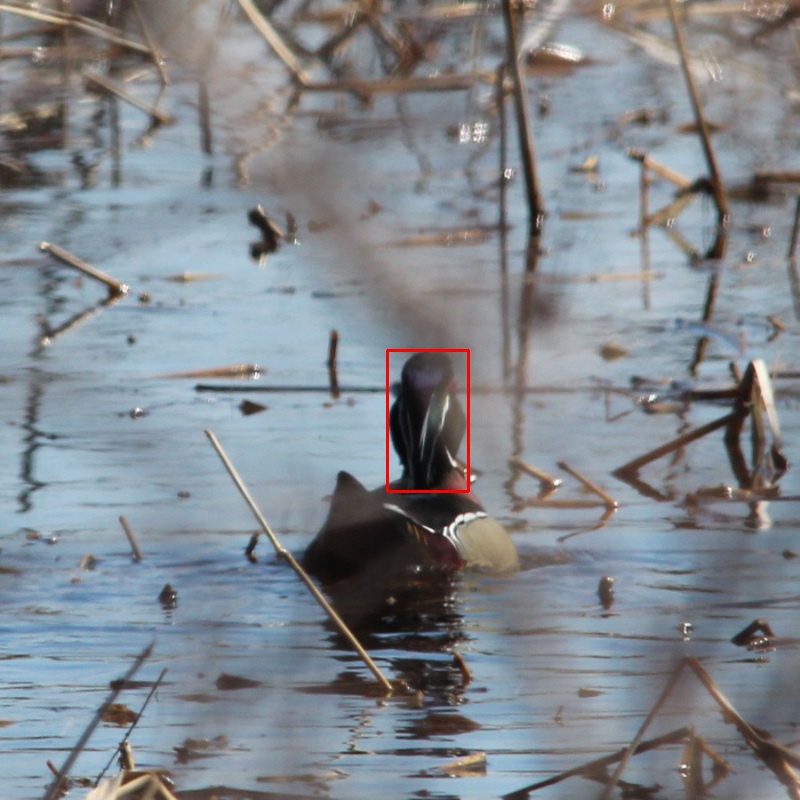}
\includegraphics[width=0.24\linewidth]{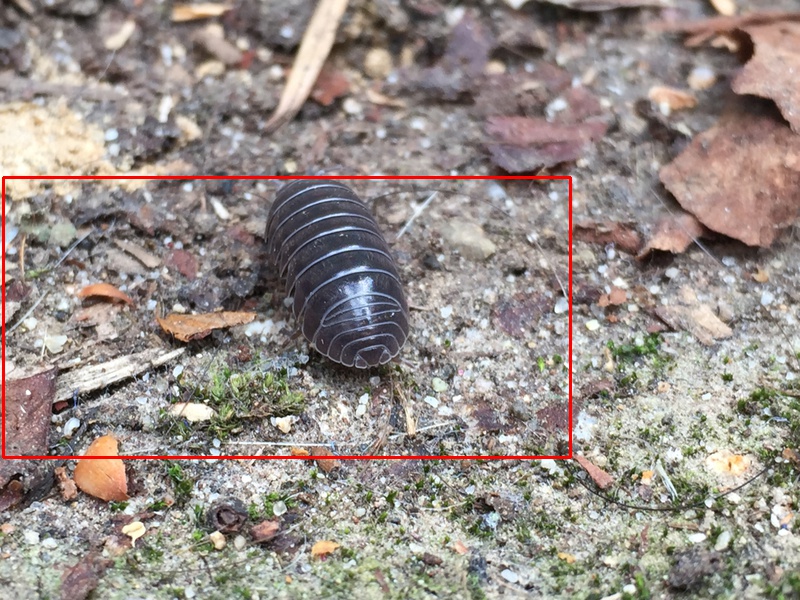}
\includegraphics[width=0.24\linewidth]{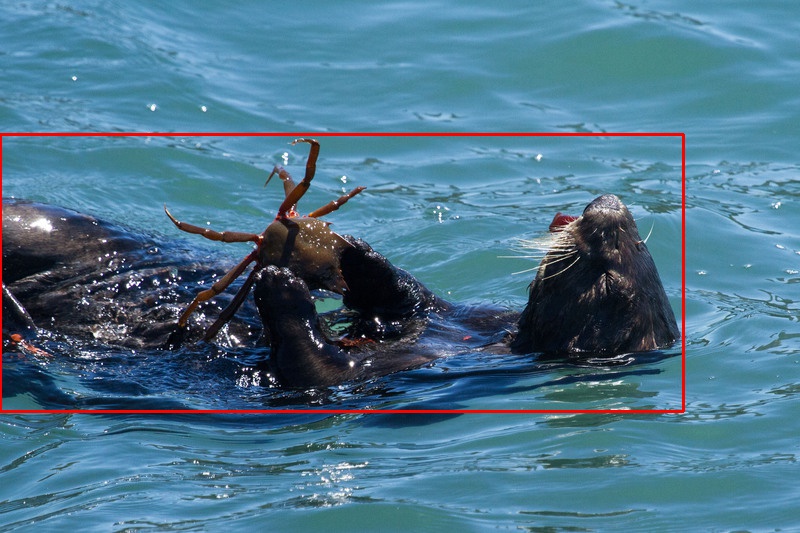}
\includegraphics[width=0.24\linewidth]{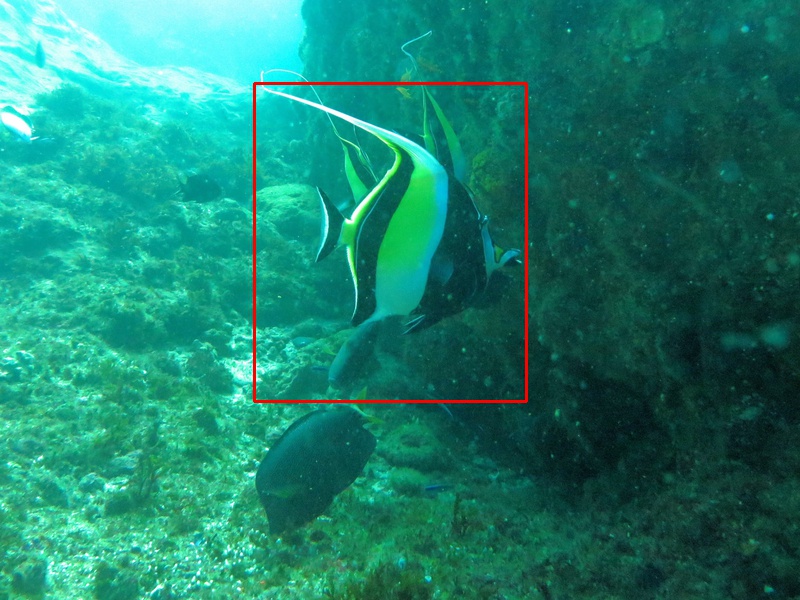} \\
\includegraphics[width=0.24\linewidth]{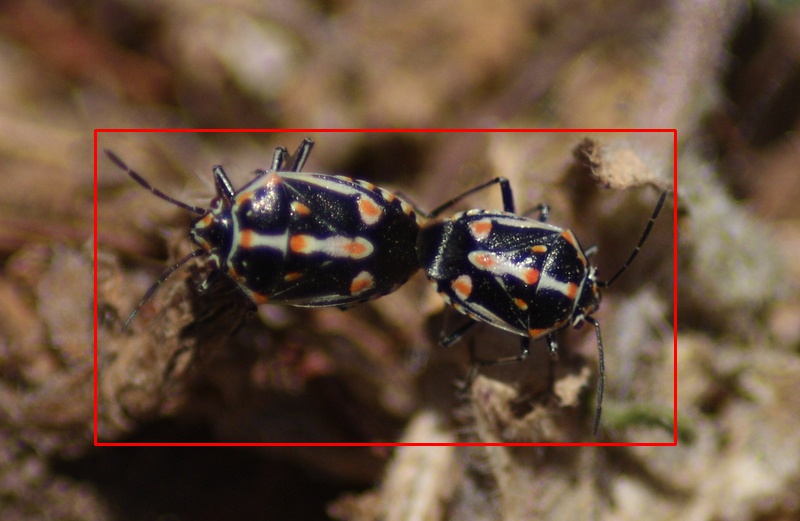}
\includegraphics[width=0.24\linewidth]{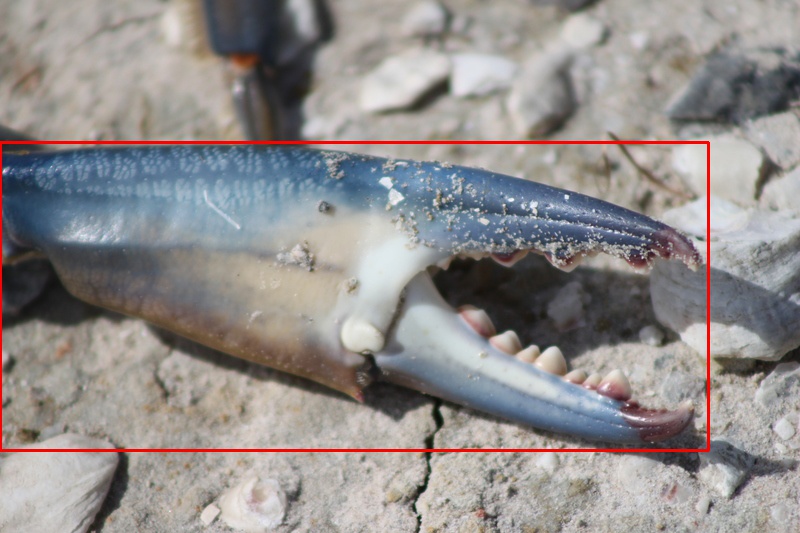}
\includegraphics[width=0.24\linewidth]{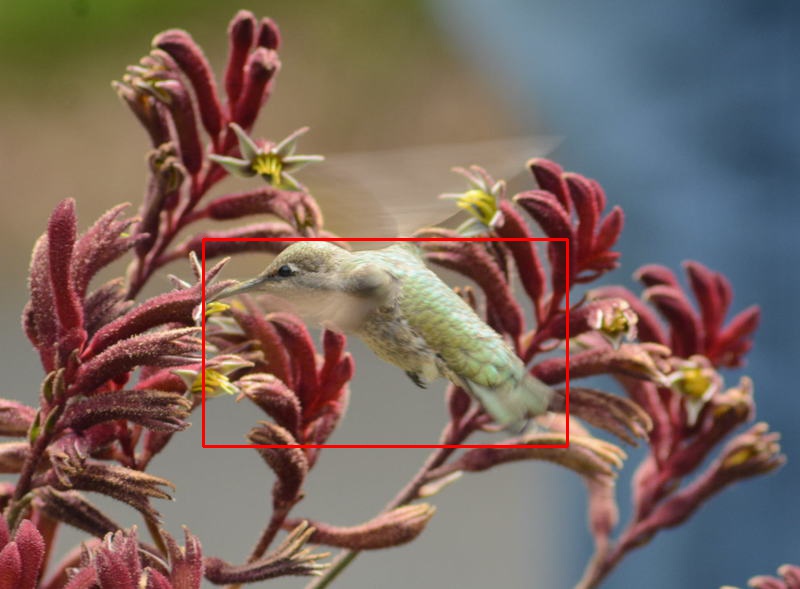}
\includegraphics[width=0.24\linewidth]{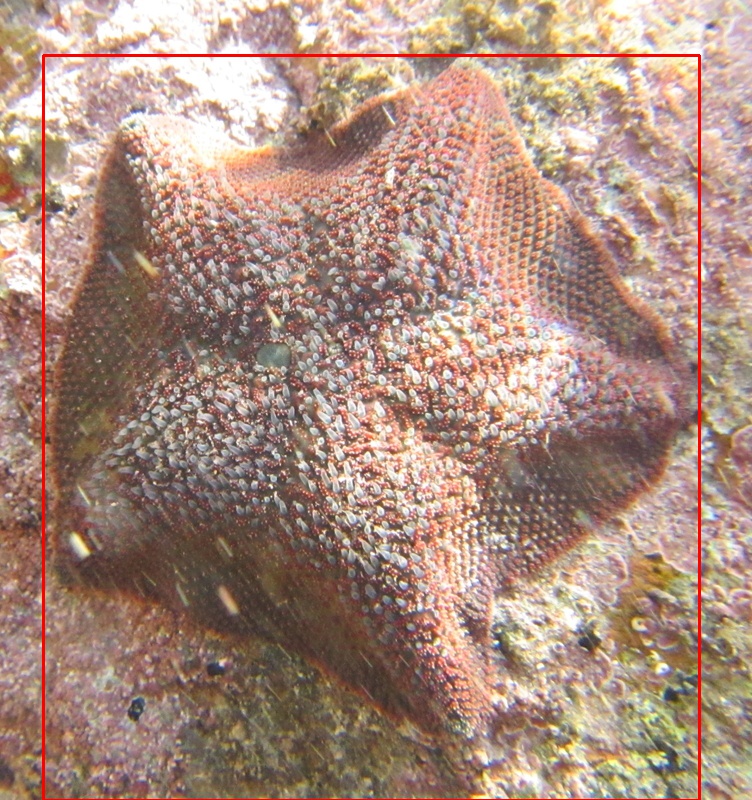}
\caption{
Examples of problematic images from iNat17 which were filtered out of the iNatLoc500 dataset. 
Each image is identified with a tuple $(i, j)$ where $i$ is the row and $j$ is the column. 
We now describe the problem in each image. 
(1, 1): The box is too small. 
(1, 2): The box is too large. 
(1, 3): The target class is the crab, not the otter, so the box is too large.
(1, 4): The box is too large and there are multiple instances of the target species. 
(2, 1): The box is too large and there are multiple instances of the target species. .
(2, 2): The image is an extreme close-up.
(2, 3): The correct box is ambiguous due to blurring. 
(2, 4): The box is too large. 
}
\label{fig:bad_examples}
\end{figure*}

We performed extensive filtering and quality control to produce \texttt{train-fullsup} and \texttt{test} splits data for iNatLoc500. We show randomly selected examples from iNatLoc500 in Fig.~\ref{fig:rand_examples}. We also show examples of images which were rejected and give the reason in each case in Fig.~\ref{fig:bad_examples}.

\section{Qualitative Examples}

\begin{figure*}[htb!]
\centering 
\includegraphics[trim={40pt 0pt 0pt 0pt},clip,height=0.9\textheight]{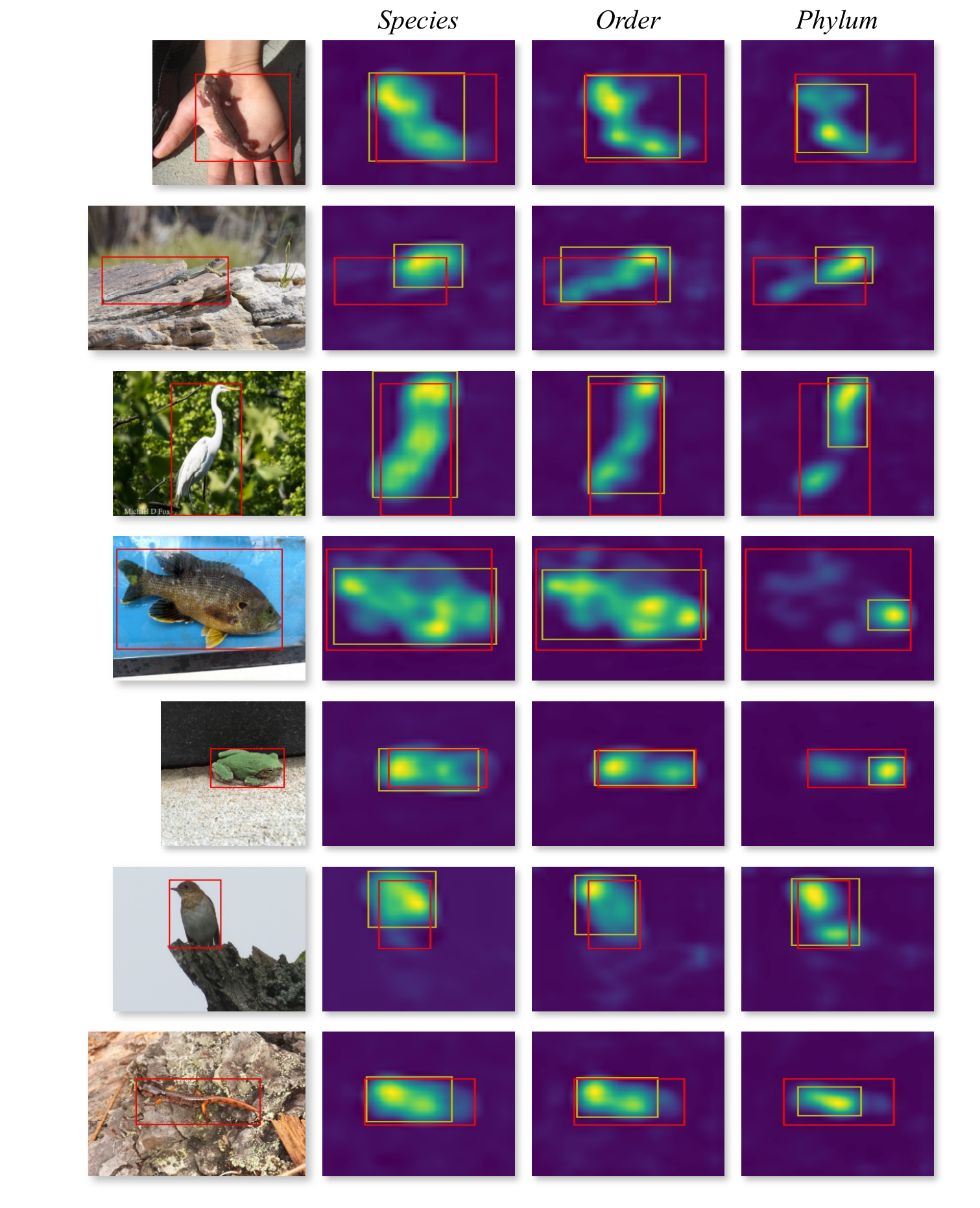}
\caption{
Examples of CAM-based WSOL predictions at different levels of granularity. In each row we provide activation map for classifiers trained at the phylum, order, and species level. Each activation map shows the ground truth bounding box (red) and WSOL-based bounding box (yellow). 
}
\label{fig:prediction_examples}
\end{figure*}

We show some hand-picked predictions for CAM-based WSOL in Fig.~\ref{fig:prediction_examples}. 

